%% file: main.tex
\documentclass[10pt,twocolumn,letterpaper]{article}

\usepackage[pagenumbers]{iccv} 

\usepackage[table,xcdraw]{xcolor}
\usepackage[accsupp]{axessibility}


\definecolor{iccvblue}{rgb}{0.21,0.49,0.74}
\usepackage[pagebackref,breaklinks,colorlinks,allcolors=iccvblue]{hyperref}
\usepackage{multirow}
\usepackage{amsmath,amssymb,amsfonts}  
\usepackage{graphicx}  
\usepackage{bbding}

\definecolor{mycolor_green}{HTML}{E6F8E0}
\definecolor{mycolor_gray}{HTML}{ECECEC}
\definecolor{mycolor_red}{HTML}{C00000}
\definecolor{mycolor_blue}{HTML}{0000CC}

\newcommand{\greenredmark}{{\color{green}{\Checkmark}}\kern-1.2ex\raisebox{1ex}{\color{red}{\rotatebox[origin=c]{125}{\textbf{--}}}}}

\def\paperID{2328} 
\def\confName{ICCV}
\def\confYear{2025}

\title{OmniEraser: Remove Objects and Their Effects in Images \\ with Paired Video-Frame Data} 
\author{Runpu Wei$^{\ast1}$
    \ \ Zijin Yin$^{\ast1}$
    \ \ Shuo Zhang$^{\ast1}$
    \ \ LanXiang Zhou$^2$
    \ \ Xueyi Wang$^1$
    \ \ Chao Ban$^2$
    \ \ \\Tianwei Cao$^1$
    \ \ Hao Sun$^2$
    \ \ Zhongjiang He$^2$
    \ \ Kongming Liang$^{\dag 1}$
    \ \ Zhanyu Ma$^1$
\\$^1$PRIS, Beijing University of Posts and Telecommunications
\\$^2$China Telecom Artificial Intelligence Technology Co.Ltd
\vspace{0.3em}
\\ \href{https://pris-cv.github.io/Omnieraser/}{\textbf{\url{https://pris-cv.github.io/Omnieraser/}}} 
}

\begin{document}

\twocolumn[{%
\renewcommand\twocolumn[1][]{#1}%
\maketitle
\vspace{-2.3em}
\begin{center}
    \centering
    \captionsetup{type=figure}
    \includegraphics[width=\textwidth]{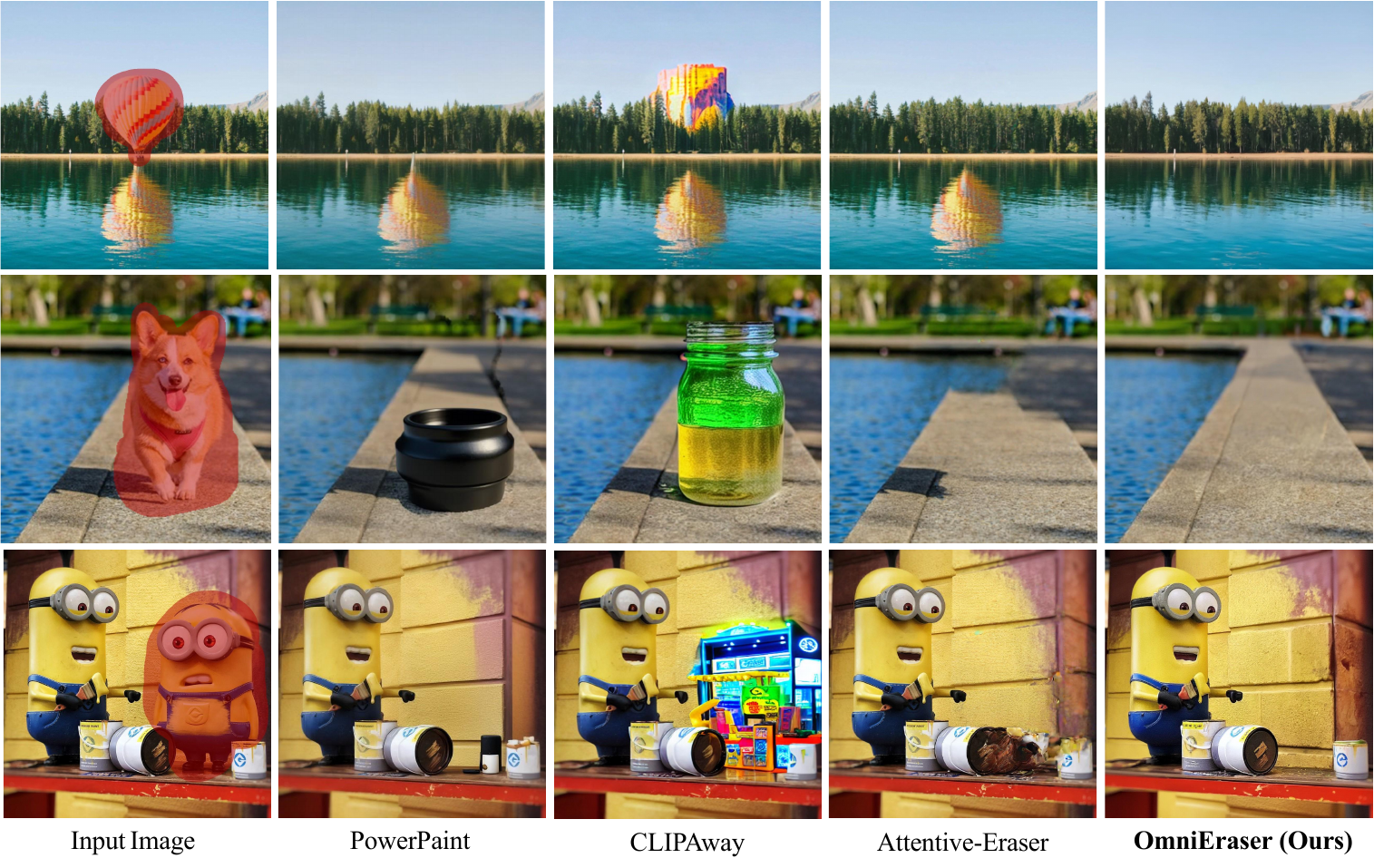}
    \vspace{-2em}
    \captionof{figure}{\textbf{OminiEraser vs. Other Methods}. State-of-the-art methods, PowerPaint \cite{zhuang2025powerpaint}, CLIPAway \cite{Ekin2024CLIPAwayHF} and Attentive-Eraser \cite{sun2024attentive}, tend to generate unintentional objects and struggle to remove the target object's effects, leading to unrealistic outputs. In contrast, our OmniEraser seamlessly removes target objects along with their shadows and reflections, using only object masks as input.}
    \label{fig:teaser}
\end{center}%
}]

\let\thefootnote\relax\footnotetext{$\ast$ Equal contribution}
\let\thefootnote\relax\footnotetext{$\dag$ Corresponding author}

\input{sec/0_abstract}
\input{sec/1_intro}

\input{sec/2_related}
\input{sec/3_method}
\input{sec/4_experiments}
\input{sec/5_conclusion}

{
    \small
    \bibliographystyle{ieeenat_fullname}
    \bibliography{main}
}

\end{document}


\maketitlesupplementary

\section{More Analysis}

\subsection{Robustness to Irregular Mask}
\begin{figure}[t]
    \centering
    \includegraphics[width=\linewidth]{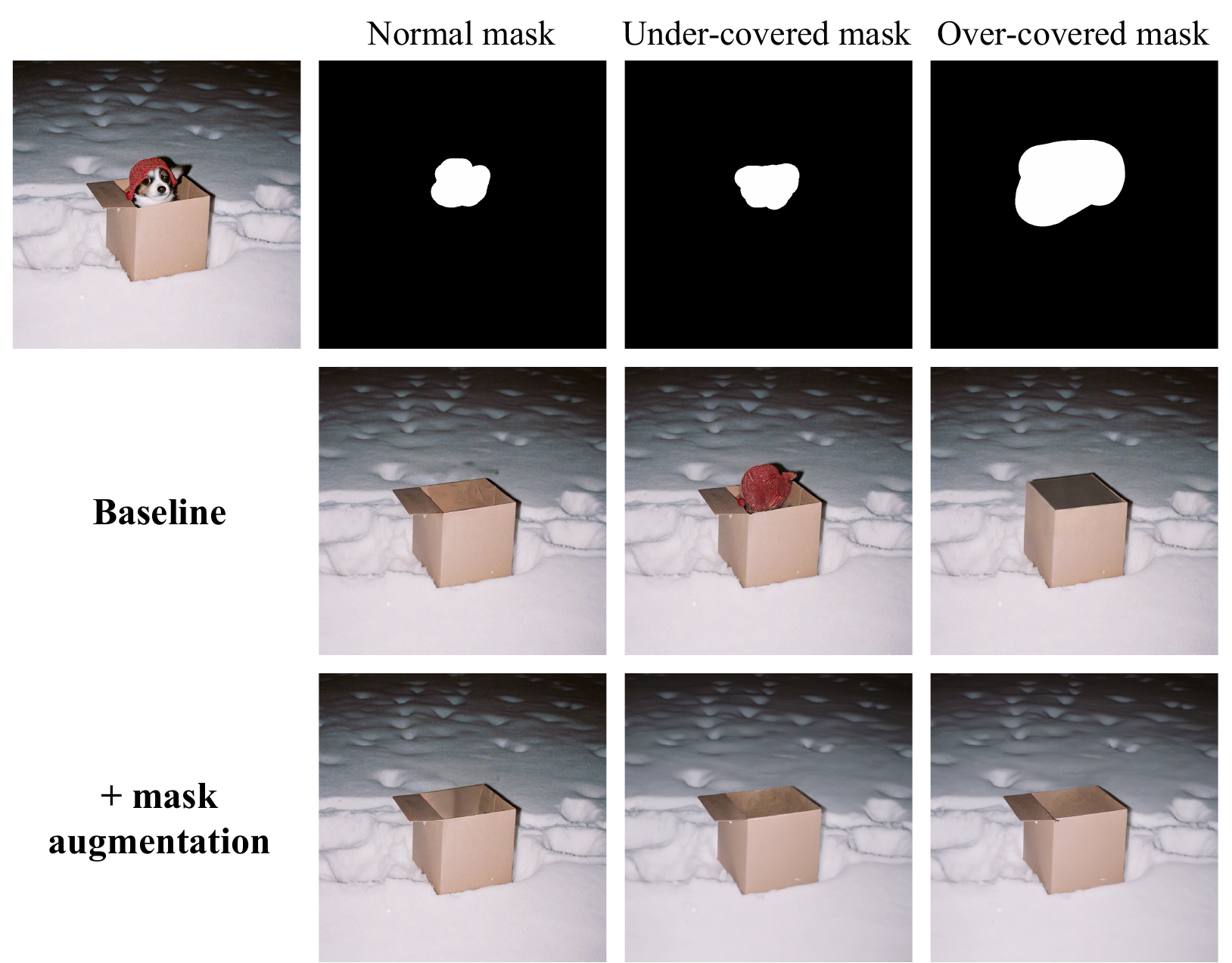}
    \caption{\textbf{Qualitative results of mask augmentation training strategy.} Our method exhibits robustness to irregular masks.}
    \label{fig:vis_mask_aug}
\end{figure}
To further demonstrate the effectiveness of our mask augmentation strategy, we qualitatively compare our method with the baseline in Fig. \ref{fig:vis_mask_aug}. We test two irregular masks: under-covered and over-covered. For normal masks, which appropriately cover the target object region, all models perform well. However, it is evident that without mask augmentation, the model tends to erase irrelevant elements and fails to remove the target object. In contrast, our method consistently removes the target object, even with irregular masks. OmniEraser demonstrates strong robustness to such irregularities, making it more user-friendly and adaptable.

\subsection{Limitation of OmniEraser}
\begin{figure}[t]
    \centering
    \includegraphics[width=\linewidth]{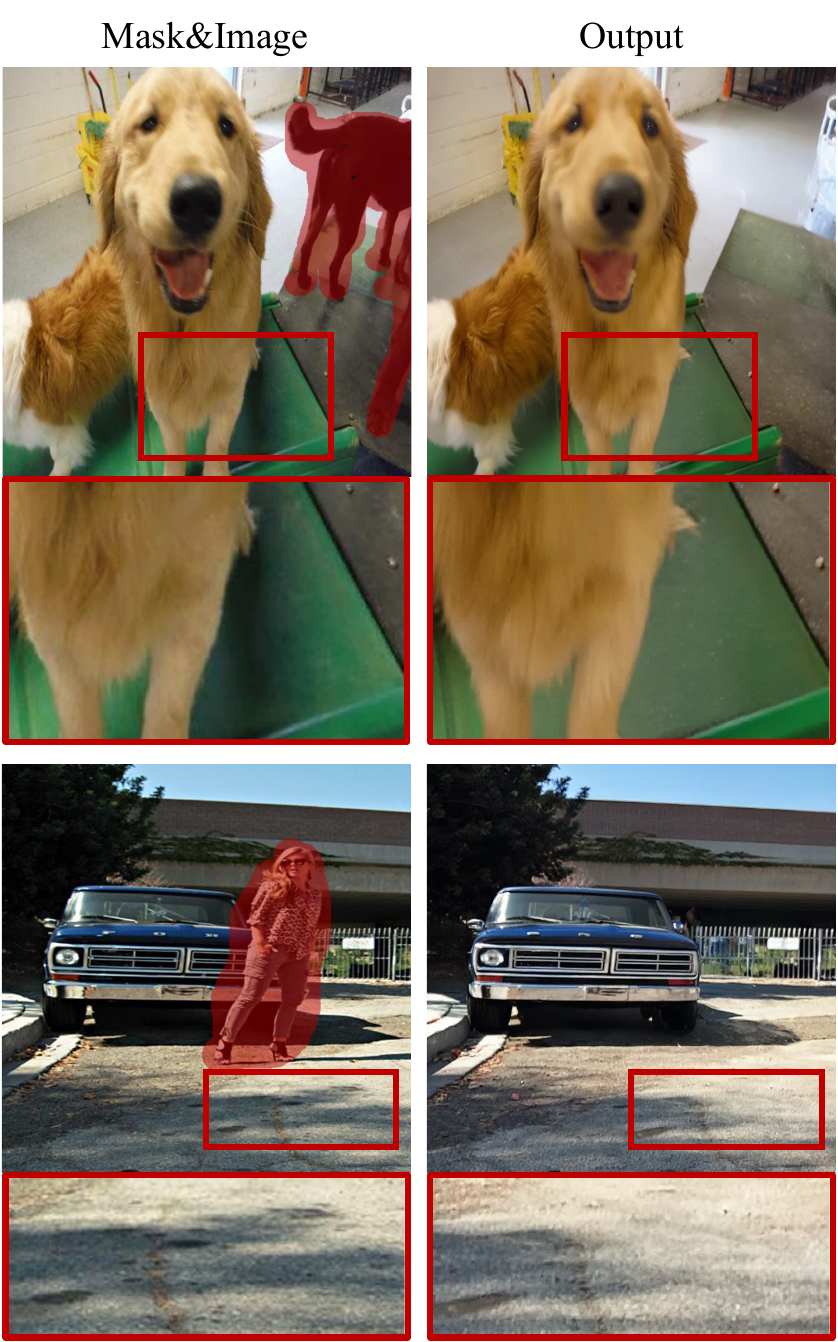}
    \caption{\textbf{Limitation of OmniEraser}. Our method could change irrelevant elements, such as other object's shadows.}
    \label{fig:vis_limitation}
    \vspace{-0.5em}
\end{figure}

\begin{figure*}[t]
\centering
\includegraphics[width=\textwidth]{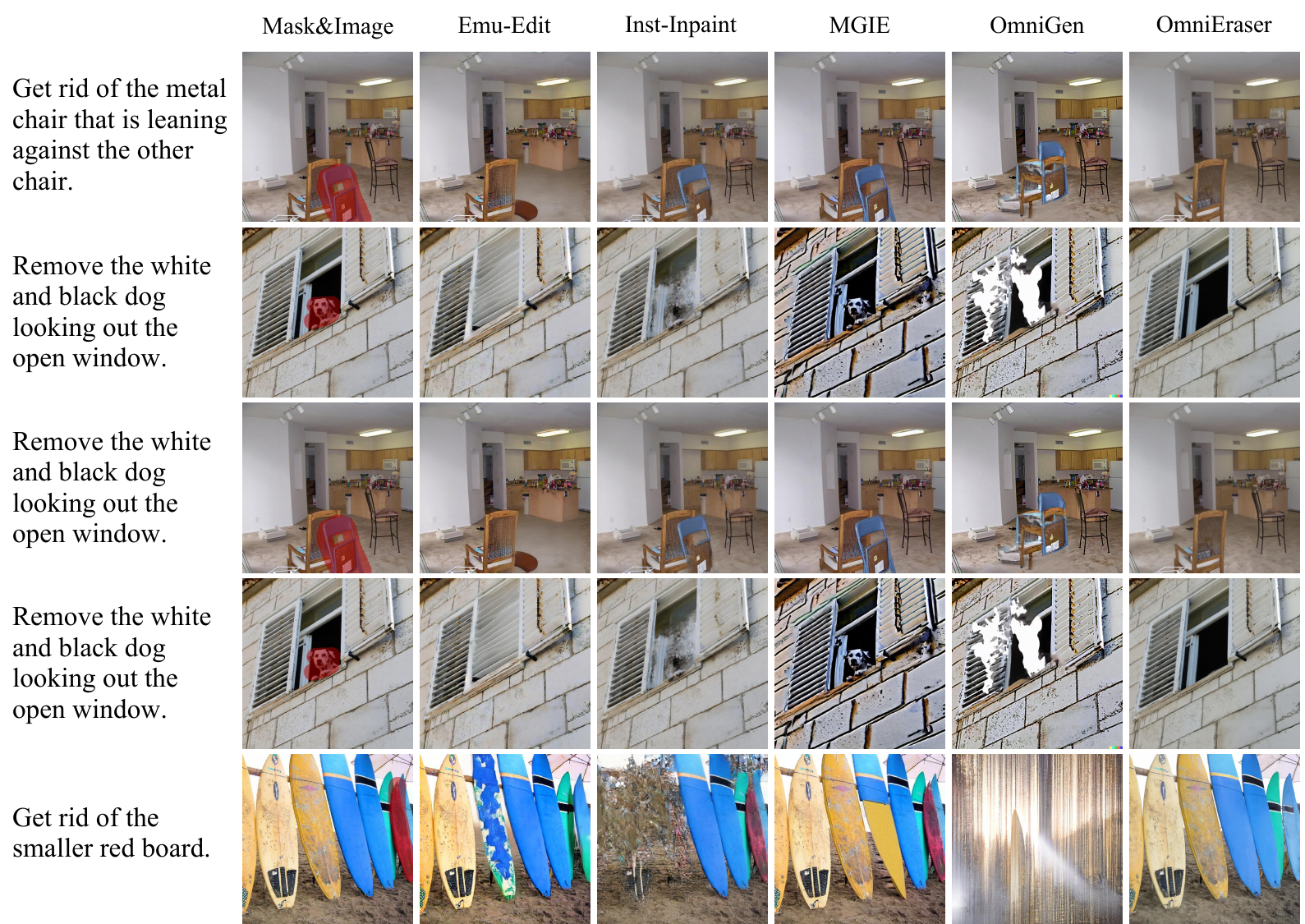}
\caption{\textbf{Comparison with existing instruction-based inpainting methods.} Our method outperforms instruction-based inpainting approaches, demonstrating superior object removal capabilities and better background preservation.}
\label{fig:vis_inst}
\end{figure*}

As observed in Fig. \ref{fig:vis_limitation}, OmniEraser may alter elements unrelated to the target object, such as shadows from other objects or background color. This issue stems from that we do not apply latent blending operation \cite{ju2024brushnet}, which fuses the foreground of the output latent with the background of the original latent, since we need to encourage the inpainting model to actively remove the object’s associated effects outside the provided mask region.

\subsection{Comparison with Instruction-based Methods.}
Previous instruction-based image inpainting methods exhibit the powerful ability to erase objects from images according to the input text. We also compare OmniEraser with these models, including Emu-Edit \cite{sheynin2024emu}, Inst-Inpaint \cite{yildirim2023inst}, MGIE \cite{fu2023guiding}, and OmniGen \cite{xiao2024omnigen}. 
Fig. \ref{fig:vis_inst} presents qualitative results using samples from the Emu-Edit test dataset \cite{sheynin2024emu}. Our method effectively removes target objects and preserves background integrity, whereas existing instruction-based approaches struggle to maintain background consistency and accurately remove the object.

\section{More Visualization Results.}
\begin{figure*}[t]
\centering
\includegraphics[width=\linewidth]{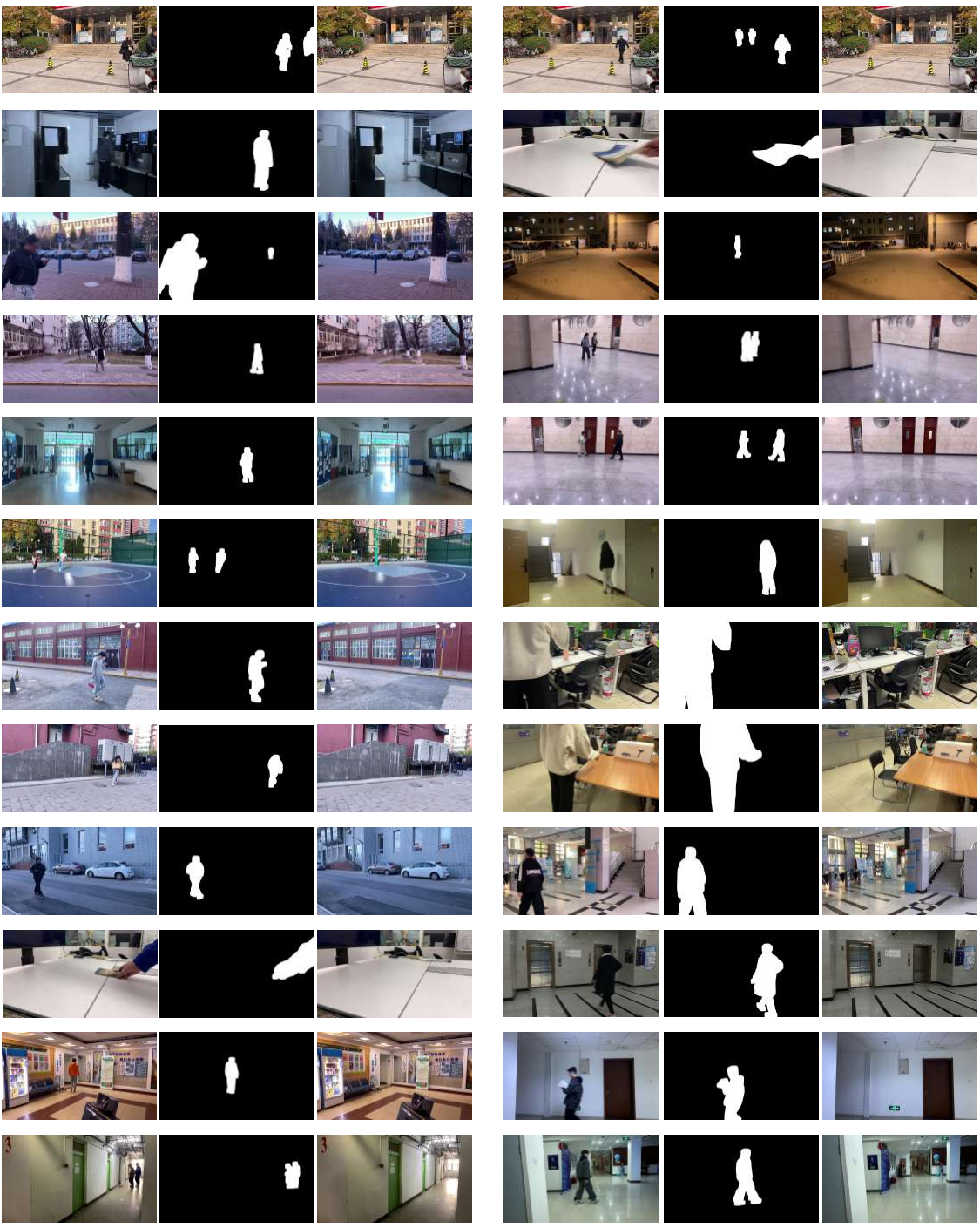}
\caption{\textbf{Examples in our \textit{Video4Removal}}. Each group of images shows the original image, the object mask, and the ground truth image.}
\label{fig:sup_vis_v4r}
\end{figure*}

\begin{figure*}[t]
\centering
\includegraphics[width=\linewidth]{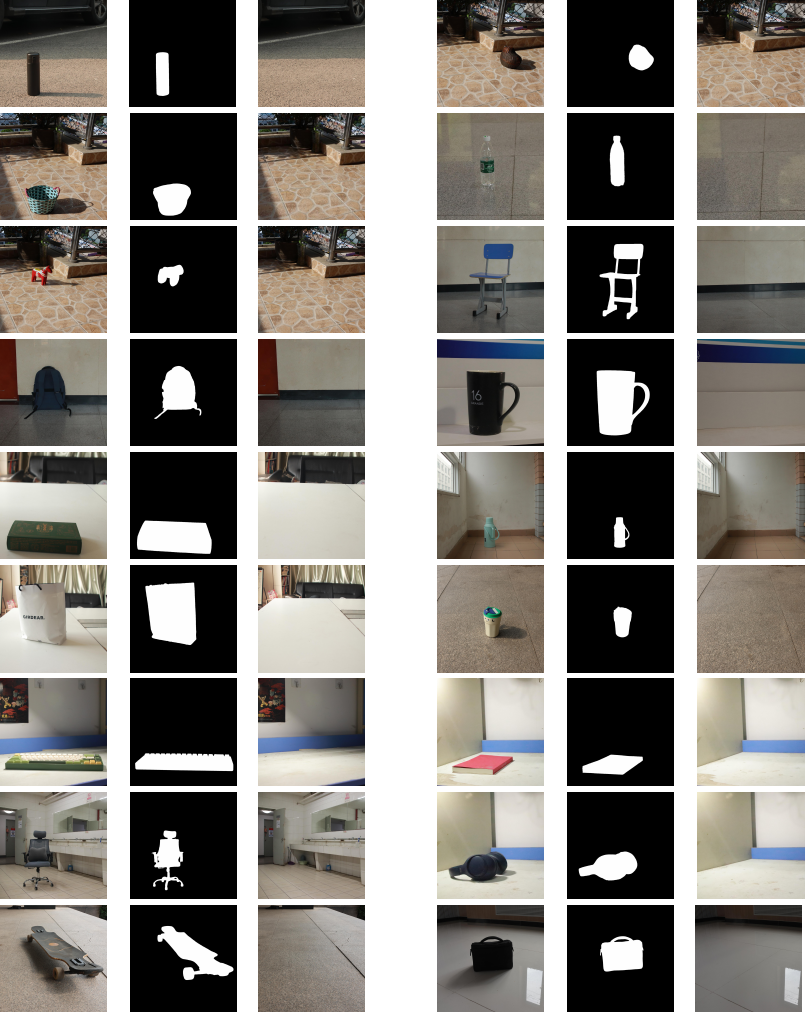}
\caption{\textbf{Examples in our \textit{RemovalBench}}. Each group of images shows the original image, the object mask, and the ground truth image.}
\label{fig:sup_vis_removalbench}
\end{figure*}

\begin{figure*}[t]
\centering
\includegraphics[width=\linewidth]{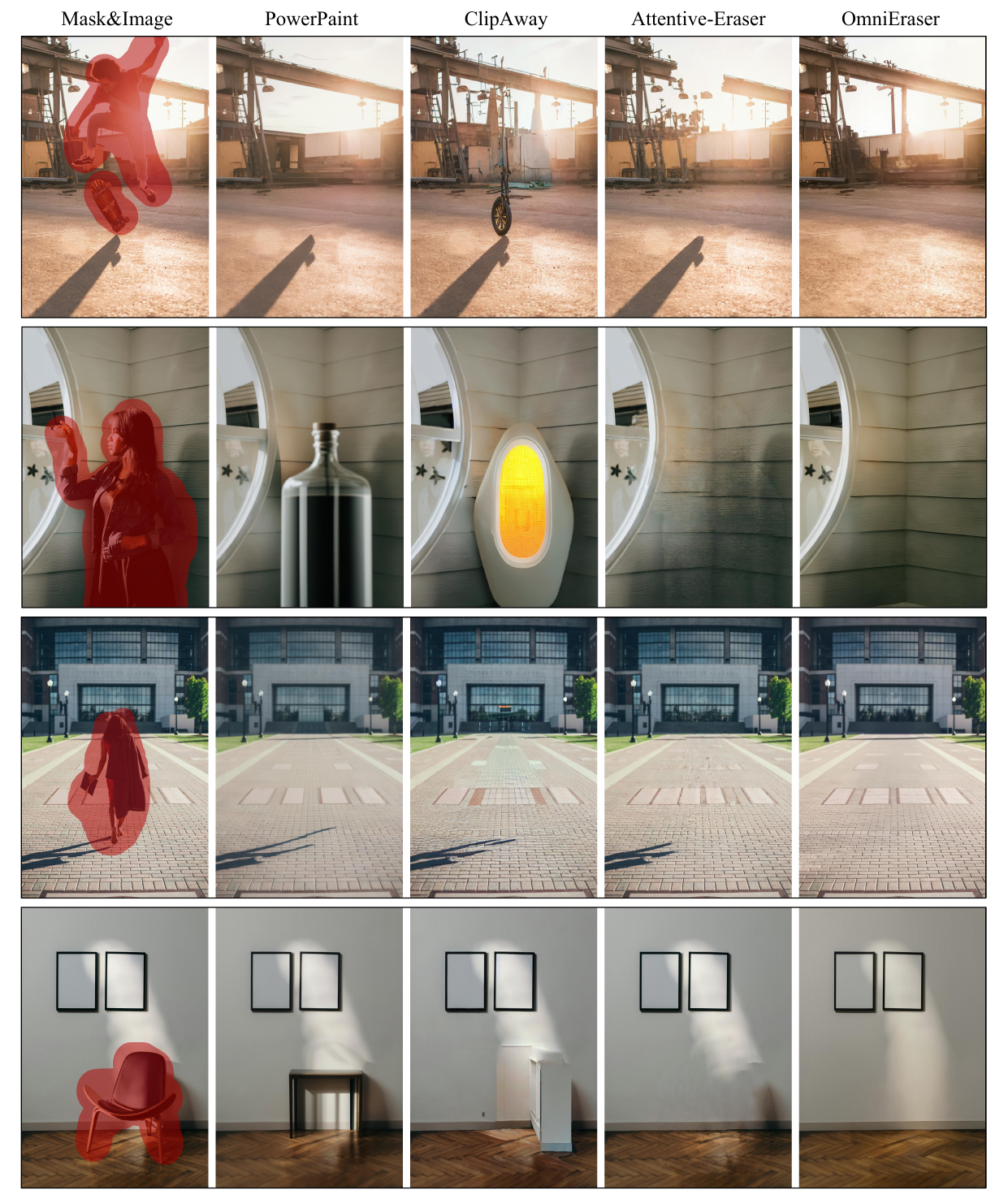}
\caption{\textbf{Qualitative comparison on in-the-wild cases.} Our OmniEraser demonstrates significant advantages.}
\label{fig:sup_vis_realwild_1}
\end{figure*}

\begin{figure*}[t]
\centering
\includegraphics[width=\linewidth]{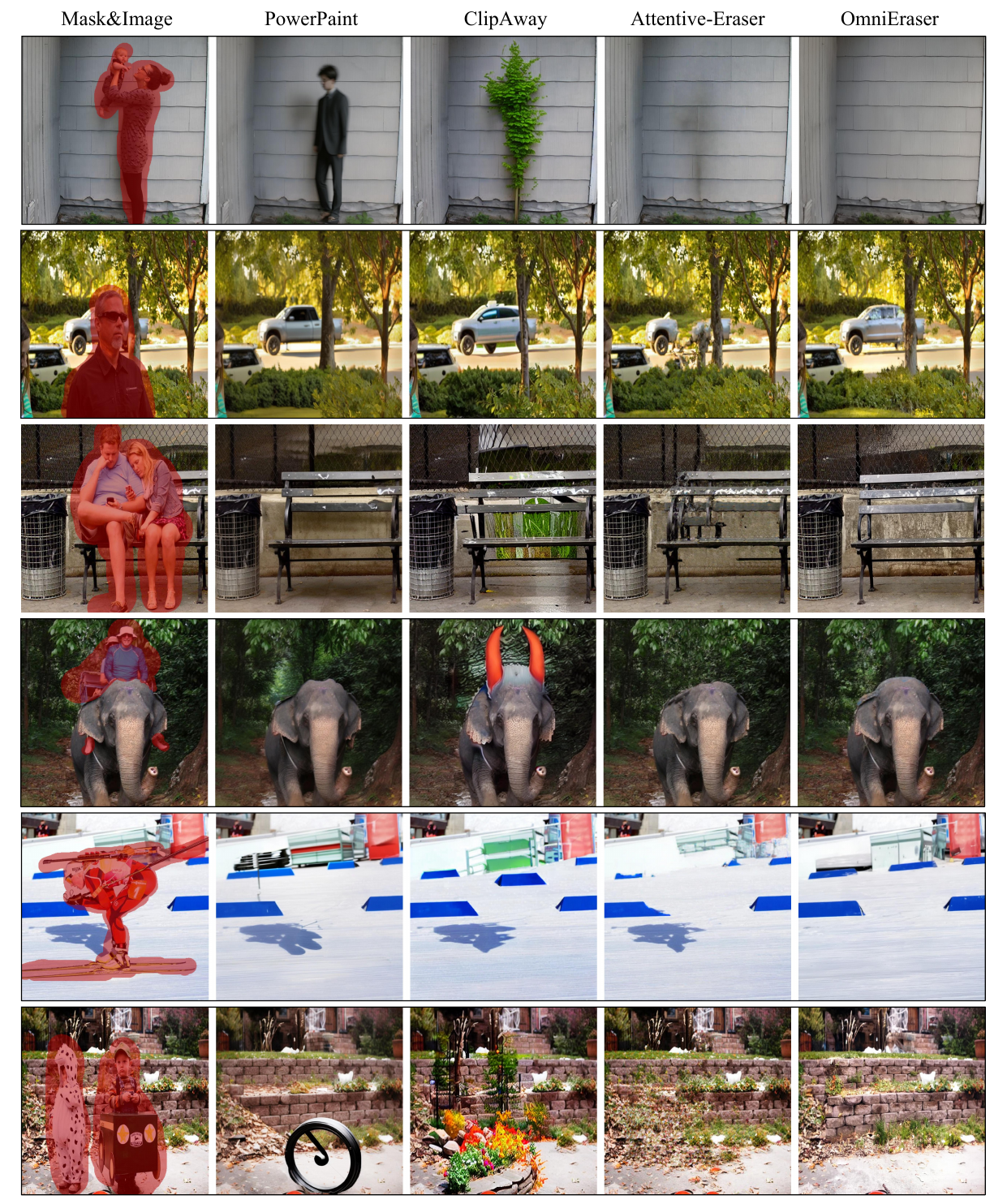}
\caption{\textbf{Qualitative comparison on in-the-wild cases.} Our OmniEraser demonstrates significant advantages.}
\label{fig:sup_vis_realwild_2}
\end{figure*}

\begin{figure*}[t]
\centering
\includegraphics[width=\linewidth]{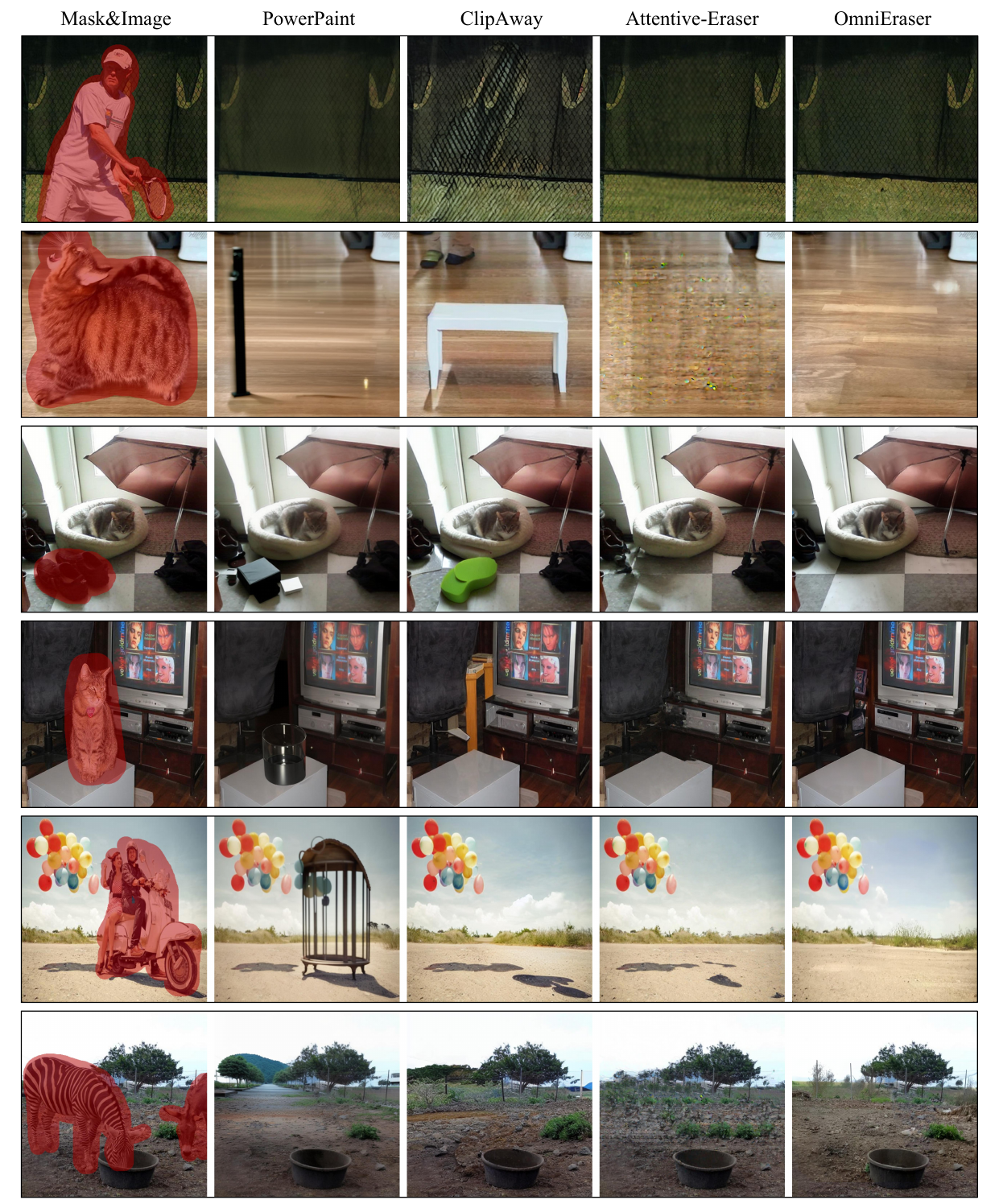}
\caption{\textbf{Qualitative comparison on in-the-wild cases.} Our OmniEraser demonstrates significant advantages.}
\label{fig:sup_vis_realwild_3}
\end{figure*}

\begin{figure*}[t]
\centering
\includegraphics[width=\linewidth]{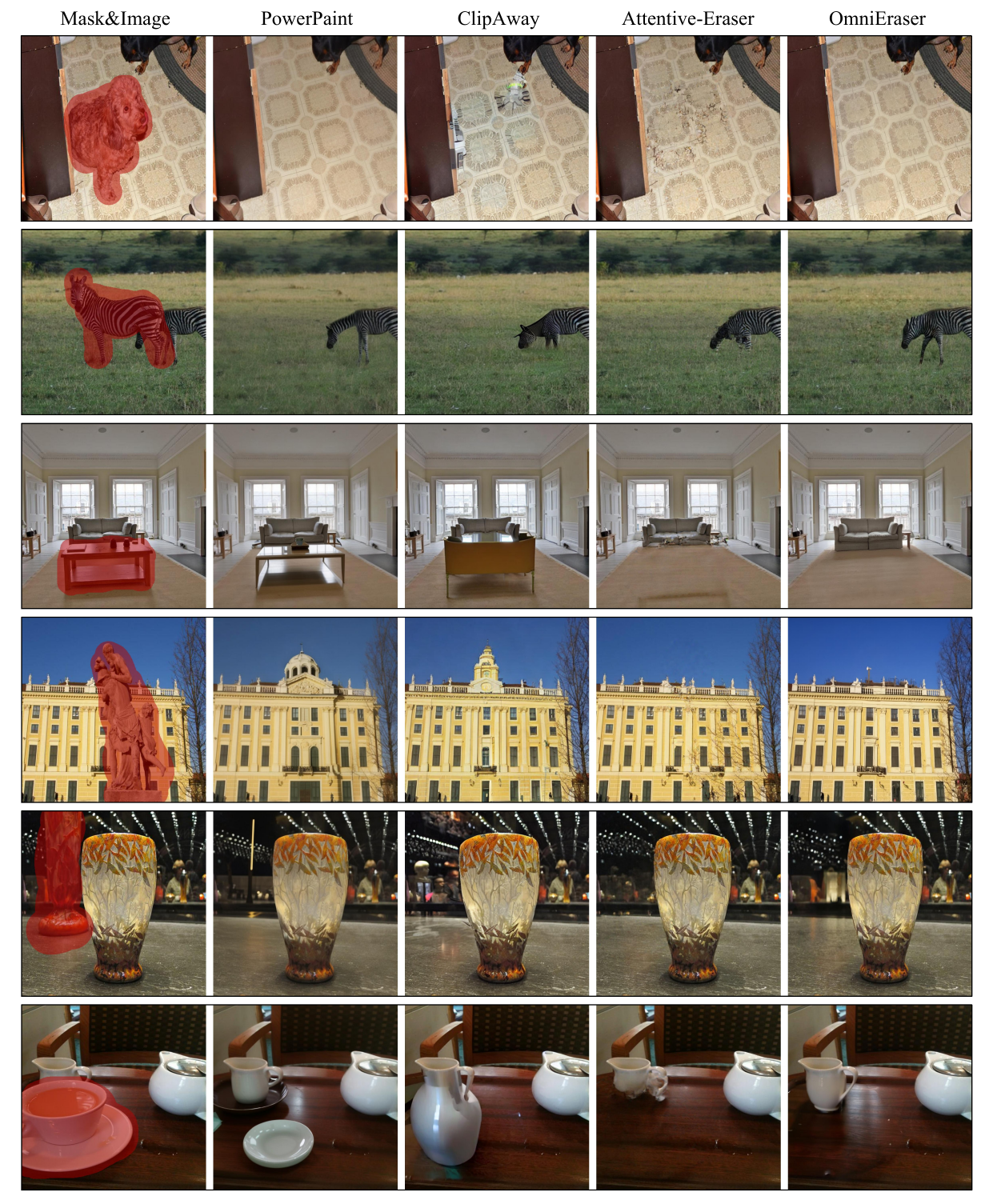}
\caption{\textbf{Qualitative comparison on in-the-wild cases.} Our OmniEraser demonstrates significant advantages.}
\label{fig:sup_vis_realwild_4}
\end{figure*}

\begin{figure*}[t]
\centering
\includegraphics[width=\linewidth]{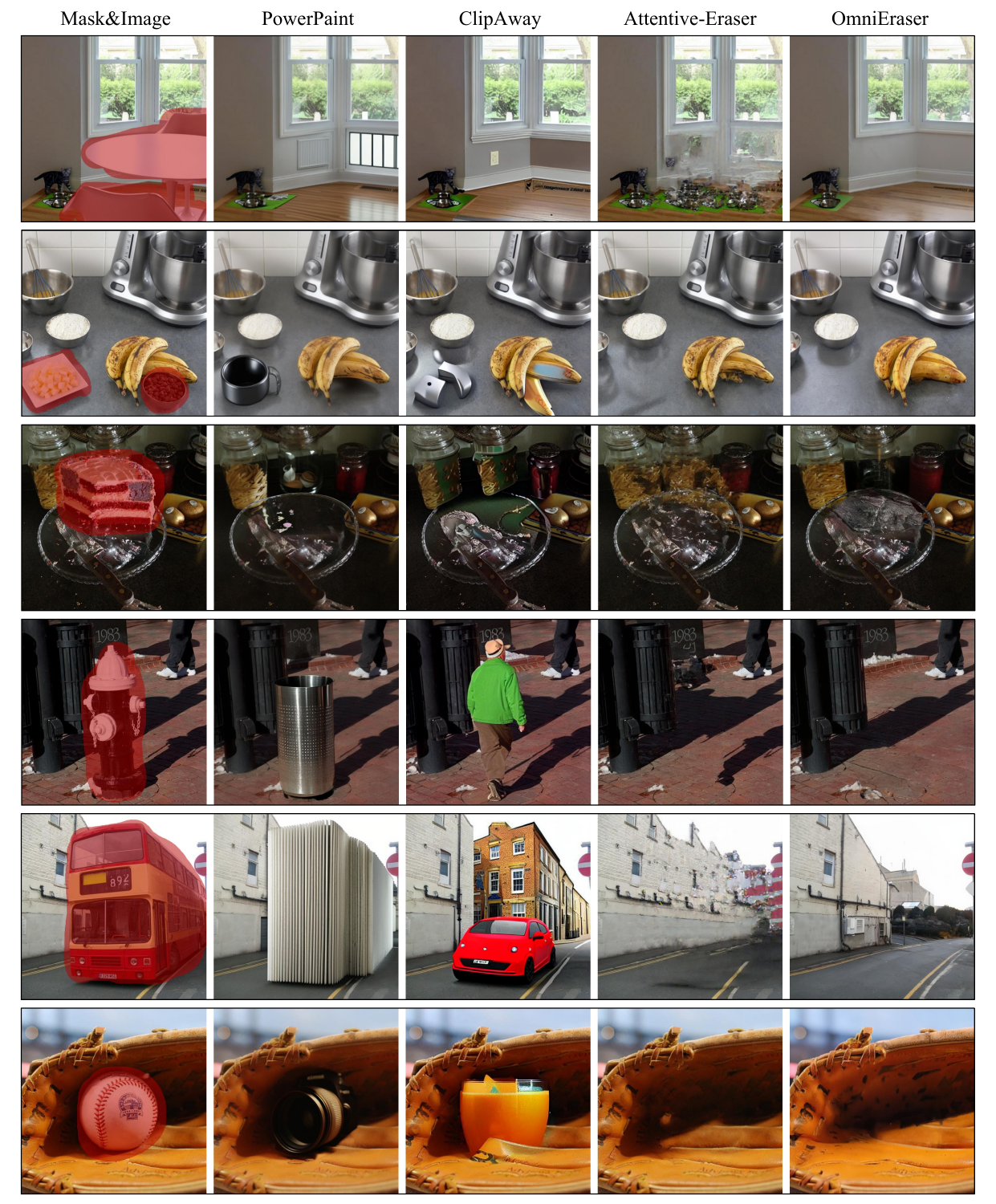}
\caption{\textbf{Qualitative comparison on in-the-wild cases.} Our OmniEraser demonstrates significant advantages.}
\label{fig:sup_vis_realwild_5}
\end{figure*}

\begin{figure*}[t]
\centering
\includegraphics[width=\linewidth]{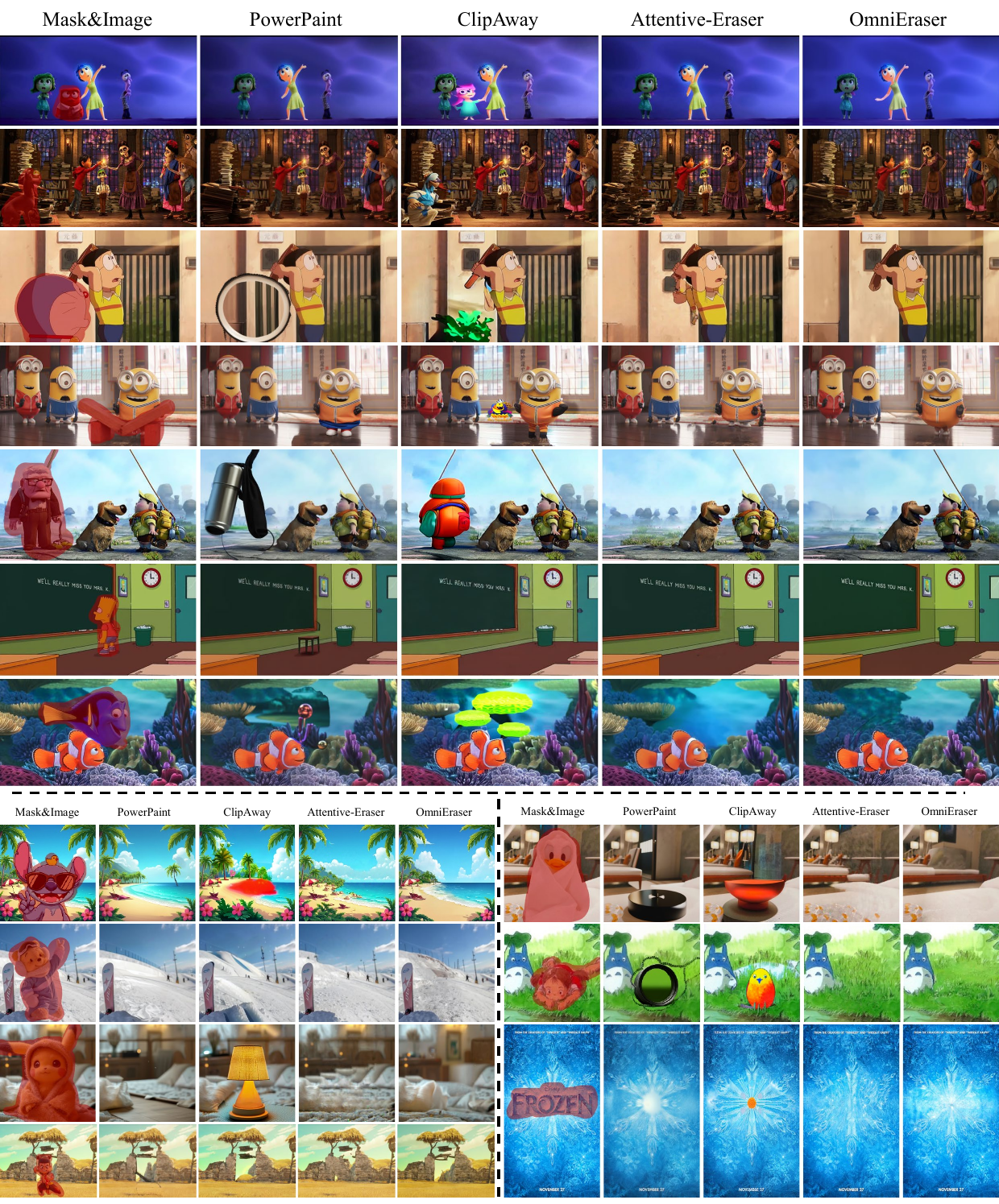}
\caption{\textbf{Qualitative comparison on anime-style images.} Our OmniEraser demonstrates significant advantages in terms of object removal quality and generalization capability across diverse anime-style images.}
\label{fig:sup_vis_cartoon}
\end{figure*}

We provide more results in this section. 
Fig. \ref{fig:sup_vis_v4r} presents additional examples from our \textbf{Video4Removal} dataset.
Fig. \ref{fig:sup_vis_removalbench} presents additional examples from our \textbf{RemovalBench} dataset.
In Fig. \ref{fig:sup_vis_realwild_1}-\ref{fig:sup_vis_realwild_5}, we provide more qualitative results in in-the-wild challenging cases. Fig. \ref{fig:sup_vis_cartoon} shows additional examples of diverse anime-style samples.
All examples collectively demonstrate OmniEraser’s strong capabilities in object removal and background preservation, and its robustness and generalization ability.

{
    \small
    \bibliographystyle{ieeenat_fullname}
    \bibliography{main}
}

%% file: sec/0_abstract.tex
\begin{abstract}

Inpainting algorithms have achieved remarkable progress in removing objects from images, yet still face two challenges: 1) struggle to handle the object's visual effects such as shadow and reflection; 2) easily generate shape-like artifacts and unintended content.
In this paper, we propose \textit{Video4Removal}, a large-scale dataset comprising over 100,000 high-quality samples with realistic object shadows and reflections. By constructing object-background pairs from video frames with off-the-shelf vision models,
the labor costs of data acquisition can be significantly reduced.
To avoid generating shape-like artifacts and unintended content, we propose Object-Background Guidance, an elaborated paradigm that takes both the foreground object and background images.
It can guide the diffusion process to harness richer contextual information.
Based on the above two designs, we present \textbf{OmniEraser}, a novel method that seamlessly removes objects and their visual effects using only object masks as input.
Extensive experiments show that OmniEraser significantly outperforms previous methods, particularly in complex in-the-wild scenes. And it also exhibits a strong generalization ability in anime-style images. 
Datasets, models, and codes will be published.

\end{abstract}

%% file: sec/1_intro.tex
\section{Introduction}
\label{sec:intro}
The rapid development of text-to-image diffusion models \cite{flux2024, Esser2024ScalingRF} has greatly improved image generation \cite{rombach2022ldm,Henschel2024StreamingT2VCD} and editing \cite{Yin2024BenchmarkingSM, brooks2023instructpix2pix} capabilities.
Among various applications, image object removal is a long-standing but challenging task, which aims to erase unwanted objects from the image and preserve the original context. 
It has broad applications, such as art design and photography post-process. 

\begin{figure}[t]
    \centering
    \includegraphics[width=\linewidth]{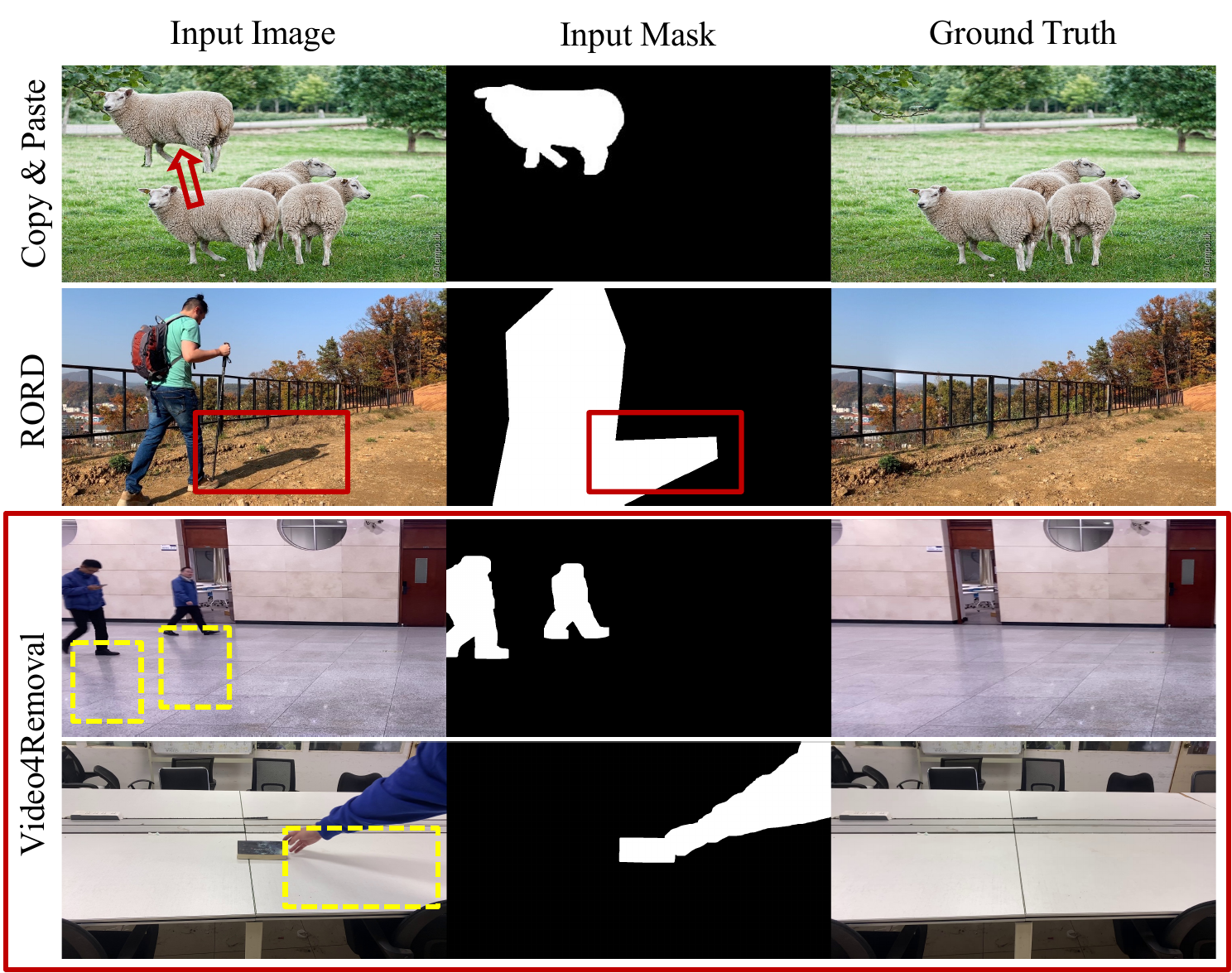}
    \caption{\textbf{Illustration of \textit{Video4Removal}.} We capture two typical visual effects: shadows and reflections. Our input masks exclude the effect regions, encouraging models to learn the associations between objects and their effects in an end-to-end manner.}
    \label{fig:video4removal}
    \vspace{-2em}
\end{figure}

Object removal requires large-scale triplet data: images with and without the target object, and the object mask. Since acquiring real data is laborious, most previous methods rely on synthesis. The common ways involve applying ``copy-and-paste’’ operation \cite{li2024magiceraser, jiang2025smarteraser} (seen Fig. \ref{fig:video4removal}), and employing off-the-shelf inpainting models to generate pseudo-labels \cite{tudosiu2024mulan}.
Although these strategies have enabled recent studies \cite{lugmayr2022repaint, ju2024brushnet, zhuang2025powerpaint, winter2024objectdrop, li2024magiceraser, li2025diffueraser, Ekin2024CLIPAwayHF} to achieve remarkable progress, they are unrealistic: lacking target objects' physical effects, \textit{i.e.} shadow and reflection. This leads to models not dealing with object shadow, reflection, and light variations in real scenarios (seen Fig. \ref{fig:teaser}). Some works have sought to construct realistic data. For example, RORD \cite{sagong2022rord} extracts real image pairs from video frames, but its annotations include shadow regions (highlighted by red lines in Fig. \ref{fig:video4removal}). This makes the annotation process laborious, and the user must manually specify the regions of object effects for removal. 
ObjectDrop \cite{winter2024objectdrop} also captures real-world samples, but it is limited in scale and not publicly available.

Moreover, previous works \cite{ju2024brushnet, zhuang2025powerpaint} often generate unintentional new objects or artifacts, such as adding a new bowl as shown in Fig. \ref{fig:teaser}. This issue arises because these methods rely solely on background context to guide the denoising process, leaving the diffusion model without explicit awareness of the target region. Consequently, the model tends to synthesize a new object based on surrounding visual cues rather than faithfully removing the target.

In this paper, we propose a new large-scale dataset \textit{Video4Removal} which contains 134,281 high-quality realistic images. Our core idea is to create triplet data from abundant video frames. We first collect videos from open-world indoor and outdoor scenes, using fixed camera viewpoints to capture the presence and movement of subjects. Then, we select frames from videos to create high-quality real pairs: scenes with and without target objects along with their effects. At last, we employ off-the-shelf models \cite{oquab2023dinov2,ravi2024sam2} to automatically generate object masks that exclude object effects' areas. Our pipeline significantly reduces labor costs and ensures dataset scalability. 
Furthermore, we propose Object-Background Guidance, which incorporates the object as an extra condition to guide the diffusion process \cite{song2020ddim, ho2020ddpm}. The core idea is to use both the foreground object image and the background image as independent guidance signals, enhancing the model’s contextual understanding of the target region and its surroundings. This paradigm ensures better background consistency while preventing the generation of unintended objects in the target region.

Built upon Object-Background Guidance, we integrate LoRA \cite{hu2022lora} into FLUX \cite{flux2024} and train it on our \textit{Video4Removal} dataset. We present \textbf{OmniEraser}, capable of removing objects and their associated effects using only the object masks. 
To ensure a more comprehensive evaluation, we provide a new benchmark \textit{RemovalBench}, which contains 70 pairs of carefully crafted objects and ground-truth examples. We also evaluate challenging in-the-wild scenes, including anime-style images. Extensive experiments demonstrate that OmniEraser achieves unprecedented performance compared to previous methods.

In summary, our key contributions are as follows:

\begin{itemize}[leftmargin=1em]
    \item We introduce \textit{Video4Removal}, a new photorealistic dataset for object removal, containing over 100,000 images derived from real-world video frames.
    \item With our proposed Object-Background Guidance, we present OmniEraser, a novel method capable of seamlessly removing both objects and their visual effects.
    \item We provide a new benchmark \textit{RemovalBench}, and extensive experiments demonstrate the superiority of \textit{OmniEraser} in terms of image quality and robustness.
\end{itemize}

%% file: sec/2_related.tex
\begin{figure*}[ht]
\centering
\includegraphics[width=\linewidth]{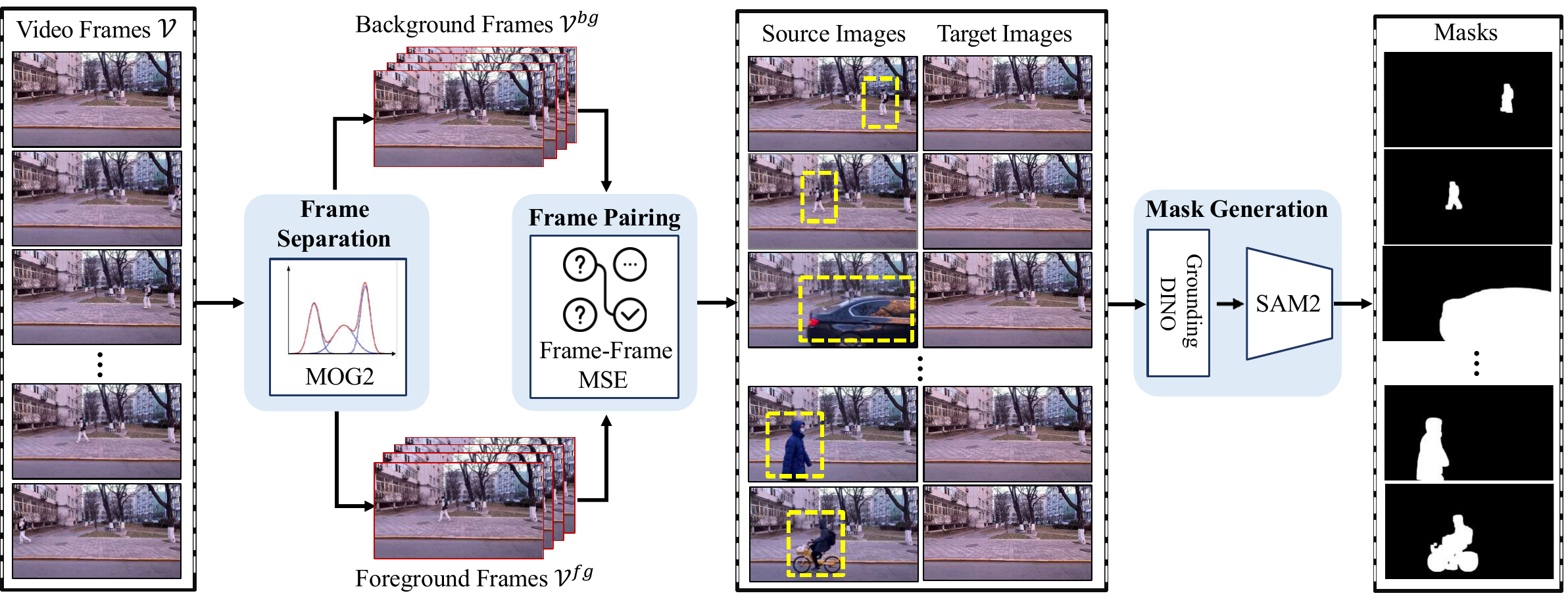}
\caption{\textbf{The construction pipeline of \textit{Video4Removal}.} We first separate all video frames into two categories: background frames and foreground frames containing moving objects. Then, each foreground frame is paired with the temporally closest background frame to form a pair. Finally, object masks for the foreground frames are obtained using off-the-shelf segmentation models. This process results in high-quality, photorealistic triplets suitable for object removal tasks.} 
\label{fig:pipeline}
\vspace{-1.5em}
\end{figure*}

\section{Related work}

\noindent\textbf{Image Inpainting.} 
Image inpainting aims to fill missing regions of an image based on the surrounding context. Traditional methods based on Generative Adversarial Networks (GANs) \cite{suvorov2022lama, dong2022zits, Cao2022ZITSII, li2022mat, zhang2022flow} are optimized to recover randomly masked regions in an image. However, these approaches are often trained separately on specific domains, such as manga \cite{xie2021manga}, face images \cite{karras2019style}, and scenes \cite{zhou2017places}, which limits their versatility in open-world applications.
Recently, the advancement of text-to-image diffusion models \cite{ramesh2022hierarchical, rombach2022high, sohl2015deep, dhariwal2021diffusion} has enabled more flexible generation based on input text prompts. For instance, Blended Latent Diffusion \cite{Avrahami2022BlendedLD}, SmartBrush \cite{xie2023smartbrush}, and SD-Inpaint \cite{rombach2022ldm} concatenate the mask and the source image to latent inputs, and generate new content for mask areas through denoising process. BrushNet \cite{ju2024brushnet} incorporates a trainable plug-and-play branch into pre-trained diffusion models for text-guided inpainting and applies blending to preserve background information. However, these approaches still generate uncertain content, and occasionally produce blurred textures. 


\noindent\textbf{Object Removal.} 
As a sub-task of image inpainting, object removal aims to eliminate target objects from images using input masks or text instructions \cite{sheynin2024emu,yildirim2023inst,fu2023guiding}. Existing methods \cite{zhuang2025powerpaint, Ekin2024CLIPAwayHF, li2024magiceraser, yildirim2023inst, winter2024objectdrop} typically fine-tune diffusion-based inpainting models on triplet data. 
The common approach is to synthesize such paired data by randomly masking image regions \cite{zhuang2025powerpaint, suvorov2022lama} or copying objects from one image and pasting them into another \cite{li2024magiceraser, jiang2025smarteraser, Ekin2024CLIPAwayHF} using existing \cite{lin2014microsoft,zhou2017places} datasets. OBJect \cite{michel2024object3dit} relies on 3D engines to render samples.
However, their data often lacks realism and essential object physical effects like shadows and reflections.
Recently, some specialized datasets \cite{sagong2022rord,winter2024objectdrop,michel2024object3dit} have emerged to provide more photorealistic object removal image pairs. 
However, RORD \cite{sagong2022rord}'s object masks cover both the objects and their associated effects, which hinders models from active learning to remove the visual effects. And ObjectDrop \cite{michel2024object3dit} is limited in size, image quality, and accessibility.
In contrast, our paired data is derived from realistic video frames, making it both large-scale and easily scalable. Additionally, our annotated masks exclude visual effect areas, thereby enabling models to adaptively remove objects and their associated effects according to the object itself geometry.

\noindent\textbf{Layer Decomposition.} 
Many studies \cite{lu2021omnimatte, lee2024generative, yang2024generative} focus on decomposing scenes into different layers, including the background and individual objects along with their effects. 
Omnimatte \cite{lu2021omnimatte} trains a U-Net \cite{ronneberger2015u} to matte regions associated with the given subject in a self-supervised manner. MULAN \cite{tudosiu2024mulan} creates an RGBA layers dataset from MS-COCO \cite{lin2014microsoft} and Laion Aesthetics 6.5 \cite{schuhmann2022laion} by leveraging off-the-shelf inpainting models. Alfie \cite{quattrini2024alfie} and LayerDiffusion \cite{zhang2024transparent} generate transparent image layers using text prompts. LayerDecomp \cite{yang2024generative} is trained on a hybrid of real-world and simulated data, enabling a wide range of applications, such as object removal, object spatial, and background editing. Our work also focuses on the relationship between objects and their effects. But we directly train a model to generate a background image instead of multiple image layers.


%% file: sec/3_method.tex
\section{Method}
\label{sec:method}
Our primary objective is to simultaneously remove both the target object and its associated visual effects from a given image, based on an input object mask. 
We introduce our work from two aspects: Sec. \ref{subsec:dataset} describes the construction of dataset \textit{Video4Removal}; Sec. \ref{subsec:method} describes our diffusion network design and fine-tuning method.

\subsection{Dataset Construction}
\label{subsec:dataset}

Object removal requires triplet training data: an input image containing the target object, a ground truth image with only the background, and an object mask. However, existing datasets often lack true physical object effects such as shadows and reflections, limiting the realism of models' results. 
To overcome this problem, we propose a new large-scale dataset \textit{Video4Removal}. It is constructed from real-world video frames captured with a viewpoint-fixed camera. This approach eliminates the need for manual collection of images, significantly reducing labor costs. Our automatic data construction pipeline is illustrated in Fig. \ref{fig:pipeline}. 

\noindent \textbf{Frame Separation.} 
We manually record both indoor and outdoor scenes using a static camera viewpoint, capturing the presence and movement of various subjects within each scene. Suppose that a video $ \mathcal{V} = \{V_1, V_2, \dots, V_T\} $, where $T$ denotes the total number of frames. 
The first step is determining whether each frame contains moving objects. In this way, we can separate all frames into foreground and background. 
While SAM2-based approaches \cite{ravi2024sam2,yuan2025sa2va} excel in video segmentation tasks, we empirically find that the traditional Mixture of Gaussian (MOG) algorithm \cite{zivkovic2004improved, zivkovic2006efficient} offers greater stability in our long video sequences, where various entities inevitably move within static background scenes. Therefore, we apply the MOG algorithm to detect moving foreground objects.

Specifically, for each frame $V_i \in \mathcal{V}$, the MOG generates a segmentation map $S_i$ as follows:
\begin{equation}
    S_i(p) = \begin{cases}
1 & \text{if pixel } p \text{ is foreground} \\
0 & \text{if pixel } p \text{ is background}
\end{cases}
\end{equation}
where $p$ indicate each pixel in frame $V_i$. Next, we compute the ratio of foreground pixels to the total number of pixels in the frame and compare it with a predefined threshold $\delta$. The frame $V_i$ is classified as a foreground frame if the ratio exceeds $\delta$, and as a background frame otherwise. We set $\delta$ to 0.15 as default. Finally, we can obtain the foreground frames $\mathcal{V}^{fg}$ and background frames $\mathcal{V}^{bg}$. 

\noindent \textbf{Frame Pairing.}
Since the background inherently changes with sun variations, we need to identify the corresponding background frame for each foreground frame $V^{fg}_{i}$ in order to construct paired input and ground truth images for object removal. The pairing process involves calculating the Mean Squared Error (MSE) between $V^{fg}_{i} \in \mathcal{V}^{bg}$ and each candidate background frame. Then the background frame with the smallest MSE value is selected as follows: 
\begin{equation}
    \arg\min_{j \in [1, N]} \text{MSE}(V^{fg}_{i}, V^{bg}_{j})
\end{equation}
where $N$ is the total number of background frames. 

\noindent \textbf{Mask Generation.}
To acquire accurate masks for objects in foreground frames, we combine two models, GroundDINO \cite{Liu2023GroundingDM,ren2024grounding} and SAM2 \cite{ravi2024sam2}. Therefore, we can obtain the training triplet $\{x, m, \hat{x}\}$, where $x$ is the foreground frame, serving as the input image with the target object, $m$ is the target object mask, $\hat{x}$ is the background frame, serving as the ground truth. Finally, we filter out instances that contain blurred objects caused by fast motion. 

This data construction pipeline significantly minimizes the labor costs associated with collecting images and annotations. Leveraging this pipeline, we construct the \textit{Video4Removal} dataset which consists of 134,281 high-quality photorealistic triplets, each with a resolution of 1920$\times$1080.

\begin{figure}[t]
\centering
\includegraphics[width=\linewidth]{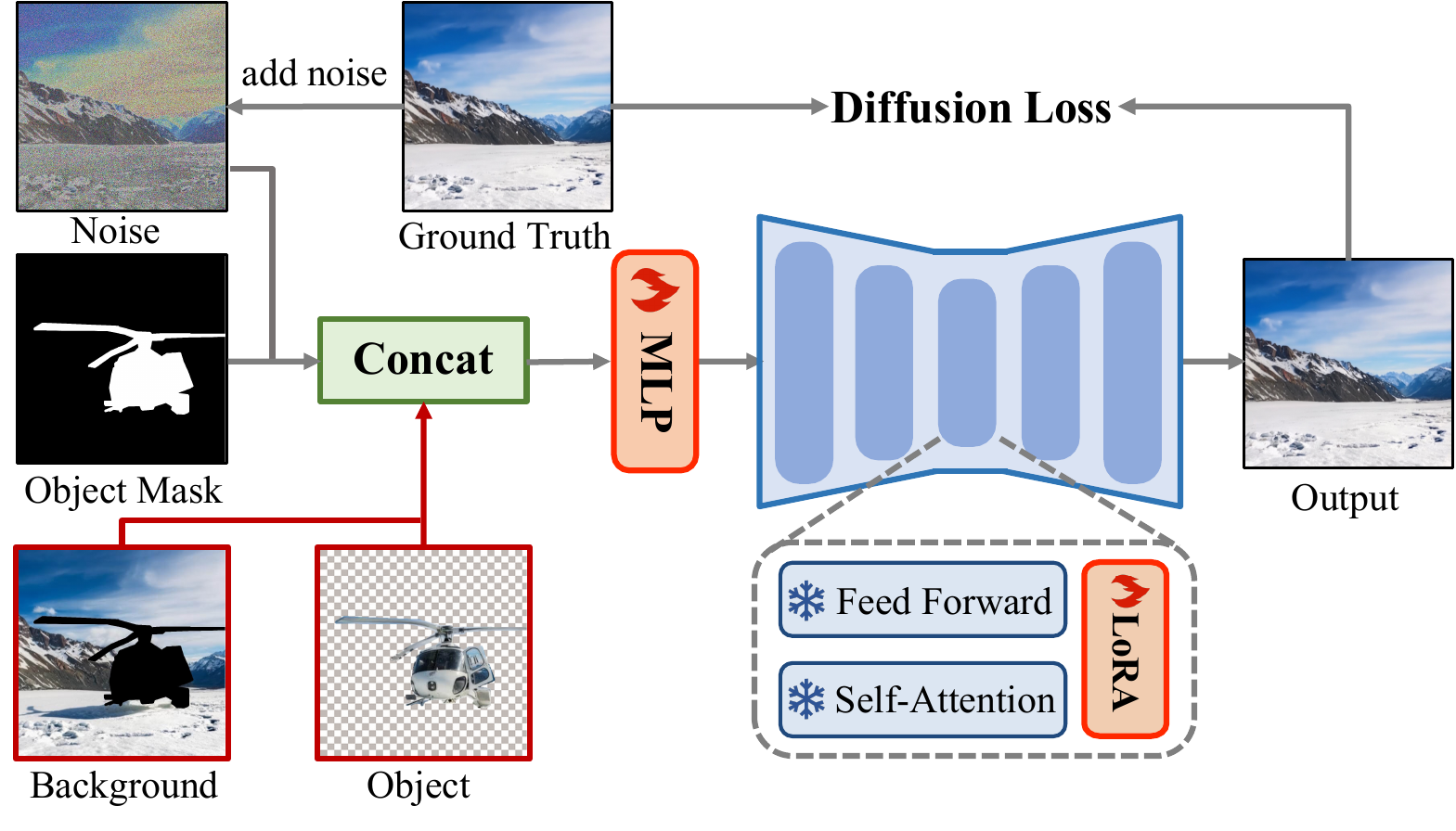}
\caption{\textbf{Architecture of OmniEraser}. We propose Object-Background Guidance which uses latent from object and background as joint input conditions to guide the denoising process.} 
\vspace{-1em}
\label{fig:method}
\end{figure}

\begin{table*}[!t]
\renewcommand\arraystretch{1}
\centering
\vspace{-0.5em}
\resizebox{\textwidth}{!}{
\begin{tabular}{lcccccccccccc}
\toprule[1.28pt]
 & \multicolumn{6}{c}{\textit{RemovalBench}} & \multicolumn{6}{c}{RORD-Val} \\ \cmidrule(lr){2-7} \cmidrule(lr){8-13}
 & FID $\downarrow$ & CMMD $\downarrow$ & LPIPS $\downarrow$ & DINO $\downarrow$ & PSNR $\uparrow$ & AS  $\uparrow$ & FID $\downarrow$ & CMMD $\downarrow$ & LPIPS $\downarrow$ & DINO $\downarrow$ & PSNR $\uparrow$ & AS  $\uparrow$ \\ 
\midrule
ZITS++ \cite{Cao2022ZITSII} & 108.38 & 0.374 & 0.158 & 0.019 & 19.62 & 4.56 & 107.44 & 0.448 & 0.274 & 0.012 & 21.17 & 4.12 \\
MAT \cite{li2022mat} & 123.78 & 0.366 & 0.164 & 0.020 & 17.88 & 4.51 & 136.53 & 0.455 & 0.281 & 0.012 & 19.18 & 4.38 \\
LaMa \cite{suvorov2022lama} & 99.88 & 0.351 & 0.156 & 0.018 & 18.72 & 4.55 & 100.21 & 0.294 & 0.229 & 0.013 & 20.50 & 4.23 \\
RePaint \cite{lugmayr2022repaint} & 102.65 & 0.741 & 0.378 & 0.018 & 19.86 & 4.38 & 114.64 & 2.345 & 0.525 & 0.016 & 17.68 & 4.71 \\
BLD \cite{Avrahami2022BlendedLD} & 128.66 & 0.553 & 0.233 & 0.022 & 17.43 & 4.39 & 224.61 & 0.862 & 0.273 & 0.015 & 17.13 & 4.74 \\
LDM \cite{rombach2022ldm} & 108.79 & 0.365 & 0.157 & 0.019 & 19.24 & 4.47 & 128.19 & 0.506 & 0.221 & 0.020 & 19.02 & 4.12 \\
SD-Inpaint \cite{rombach2022ldm} & 119.60 & 0.419 & 0.274 & 0.025 & 17.02 & 4.48 & 143.69 & 0.494 & 0.308 & 0.013 & 16.83 & 4.61 \\
SDXL-Inpaint \cite{rombach2022ldm} & 104.97 & 0.398 & 0.187 & 0.019 & 17.87 & \textcolor{mycolor_blue}{\underline{4.63}} & 147.01 & 0.460 & 0.210 & 0.012 & 17.69 & 4.76 \\
BrushNet \cite{ju2024brushnet} & 120.97 & 0.549 & 0.191 & 0.022 & 18.68 & \textcolor{mycolor_blue}{\underline{4.63}} & 234.87 & 0.745 & 0.293 & 0.031 & 16.51 & 4.41 \\
FLUX.1-Fill \cite{flux2024} & 115.79 & 0.487 & 0.193 & 0.021 & 17.12 & 4.59 & 141.39 & 0.450 & 0.217 & 0.011 & 18.50 & 4.55 \\
PowerPaint \cite{zhuang2025powerpaint} & 114.55 & 0.392 & 0.240 & 0.019 & 18.25 & 4.56 & 102.33 & 0.408 & 0.241 & 0.009 & 18.29 & 4.38 \\
CLIPAway \cite{Ekin2024CLIPAwayHF} & 108.40 & 0.272 & 0.254 & 0.015 & 18.78 & 4.48 & 81.28 & 0.545 & 0.278 & 0.011 & 16.36 & 4.19 \\
Attentive-Eraser \cite{sun2024attentive} & \textcolor{mycolor_blue}{\underline{55.49}} & \textcolor{mycolor_blue}{\underline{0.232}} & \textcolor{mycolor_blue}{\underline{0.146}} & \textcolor{mycolor_blue}{\underline{0.009}} & 20.60 & 4.50 & 96.77 & 0.233 & 0.221 & 0.008 & 20.24 & \textcolor{mycolor_blue}{\underline{4.77}} \\
\midrule
\rowcolor{mycolor_green} OmniEraser & \textcolor{mycolor_red}{\textbf{39.52}} & \textcolor{mycolor_red}{\textbf{0.208}} & \textcolor{mycolor_red}{\textbf{0.133}} & \textcolor{mycolor_red}{\textbf{0.007}} & \textcolor{mycolor_red}{\textbf{21.11}} & \textcolor{mycolor_red}{\textbf{4.66}} & \textcolor{mycolor_red}{\textbf{43.71}} & \textcolor{mycolor_red}{\textbf{0.153}} & \textcolor{mycolor_red}{\textbf{0.166}} & \textcolor{mycolor_red}{\textbf{0.006}} & \textcolor{mycolor_red}{\textbf{22.13}} & \textcolor{mycolor_red}{\textbf{4.99}} \\

\rowcolor{mycolor_green} OmniEraser (SDXL) & 86.26 & 0.348 & 0.217 & 0.014 & 20.44 & 4.52 & 46.42 & 0.249 & \textcolor{mycolor_blue}{\underline{0.190}} & 0.008 & 19.49 & 4.50 \\
\rowcolor{mycolor_green} OmniEraser (SD3) & 72.89 & 0.365 & 0.169 & 0.014 & \textcolor{mycolor_blue}{\underline{21.07}} & 4.59 & \textcolor{mycolor_blue}{\underline{44.88}} & \textcolor{mycolor_blue}{\underline{0.200}} & 0.225 & \textcolor{mycolor_blue}{\underline{0.007}} & \textcolor{mycolor_blue}{\underline{21.30}} & 4.66 \\

\bottomrule[1.28pt]
\end{tabular}}

\caption{\textbf{Quantitative results of all comparison methods.} \textcolor{mycolor_red}{\textbf{Red}} and \textcolor{mycolor_blue}{\underline{blue}} colors represent the best and second best performance. Our OmniEraser achieves the best performance across all metrics. Notably, using other alternative base models, \textit{i.e.} SDXL and SD3, OmniEraser still delivers competitive results.}
\label{tab:main_bench}

\end{table*}

\subsection{Model Design}
\label{subsec:method}
We train the diffusion model using \textit{Video4Removal} with the standard diffusion loss, as shown in Fig. \ref{fig:method}. Below, we introduce our key designs.

\noindent \textbf{Generative Prior.}
Several large-scale diffusion models, including FLUX.1-dev \cite{flux2024} and SD-XL \cite{rombach2022ldm}, have demonstrated remarkable performance in text-to-image generation. For our task, we choose FLUX.1-dev \cite{flux2024}, which utilizes the DiT network \cite{peebles2023scalable}, as our generative prior.

\noindent \textbf{Object-Background Guidance.}
Previous inpainting methods \cite{zhuang2025powerpaint, ju2024brushnet, Li2025RORemTA} disregard the masked region in their model inputs, suffering from generating a new shape-like object in the target area. Recently, SmartEraser \cite{jiang2025smarteraser} uses the full image as guidance for the denoising process to provide more contextual information. 
In contrast, we propose the Object-Background Guidance, in which the foreground object and background images serve as independent input guidance. Specifically, we concatenate the latent representations of the masked object, masked background, noise, and the mask itself, then pass them through a trainable linear layer. 
This enhances the model’s understanding of the target region and background context, enabling better background consistency and preventing the generation of a new object. 

\noindent \textbf{Adaptor.} 
Low-Rank Adaptation (LoRA)  \cite{hu2022lora} is an effective method designed to adapt large-scale models. Thus, we integrate LoRA \cite{hu2022lora} into both the self-attention and feed-forward layers of each DiT block, keeping the parameters of the base model frozen during training.


\noindent \textbf{Mask Augmentation.}
To enhance the method's robustness to irregular masks that users may provide in inference, we increase the diversity of input masks in terms of shape and size during training. Following \cite{suvorov2022lama, jiang2025smarteraser}, we adopt the following morphological strategy: (1) dilation: expanding the object boundaries, (2) erosion: shrinking the object boundaries, (3) box: generating the smallest enclosed rectangular box around the original mask. During training, for each input mask, we randomly apply one of these augmentations or leave the mask unchanged.

%% file: sec/4_experiments.tex
\section{Experiments}
\label{sec:experiments}
In this section, we conduct extensive experiments to evaluate and analyze our OmniEraser. Sec. \ref{subsec:exp_set} describes experimental settings, Sec. \ref{subsec:exp_compare} compares OmniEraser with existing methods, and Sec. \ref{subsec:exp_ablation} conducts ablation studies.

\subsection{Setting}
\label{subsec:exp_set}
\noindent \textbf{Implementation Details. }
We use the FLUX.1-dev model \cite{flux2024} as generative prior, and the LoRA adapter rank is set to 32. The batch size is 1, the learning rate is 5e-5, and the guidance scale is 3.5. We use the text prompt ``There is nothing here'' for inference and training. Our OmniEraser and all ablation models are trained for 130,000 steps on a single NVIDIA Tesla A800 GPU, which takes around 1 day. All experimental results are obtained by 28 denoising steps. More details can be found in our open-source code.

\noindent \textbf{Comparison Methods. }
We compare our approach with state-of-the-art methods in both inpainting and object removal. For inpainting, we evaluate traditional GAN-based methods, such as ZITS++ \cite{Cao2022ZITSII}, MAT \cite{li2022mat}, LaMa \cite{suvorov2022lama}, and diffusion-based methods, including RePaint \cite{lugmayr2022repaint}, BLD \cite{Avrahami2022BlendedLD}, LDM \cite{rombach2022ldm}, and BrushNet \cite{ju2024brushnet}. Additionally, we compare with open-source image inpainting suites like SD-Inpaint \cite{rombach2022ldm}, SDXL-Inpaint \cite{rombach2022ldm}, and FLUX.1-Fill \cite{flux2024}. For specialized object removal, we consider PowerPaint \cite{zhuang2025powerpaint}, CLIPAway \cite{Ekin2024CLIPAwayHF}, and Attentive-Eraser \cite{sun2024attentive}. For all models, we use their official weights and default recipes.

\begin{figure}[t]
    \centering
    \includegraphics[width=\linewidth]{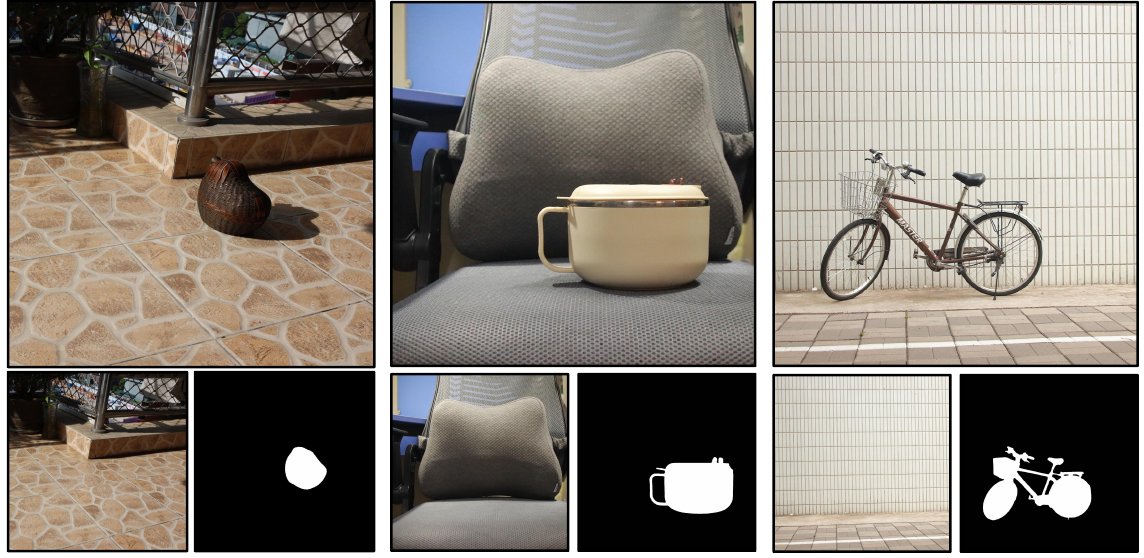}
    \caption{\textbf{Examples from \textit{RemovalBench}}. Each group of images shows the original image, ground truth, and object mask.}
    \vspace{-1.5em}
    \label{fig:removalbench}
\end{figure}

\begin{figure*}[t]
    \centering
    \includegraphics[width=\textwidth]{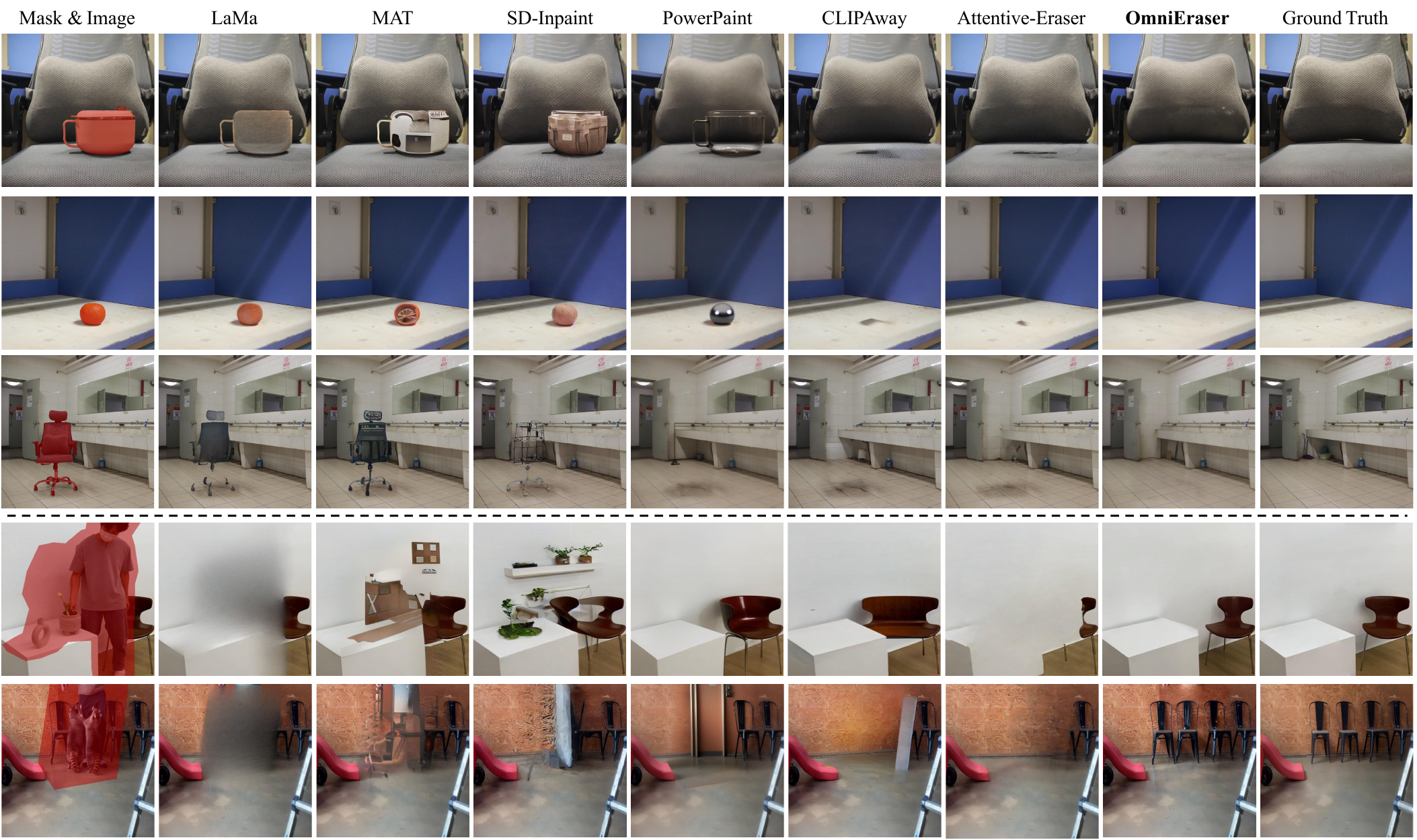}
    \caption{\textbf{Qualitative results of all comparison methods on \textit{RemovalBench} (the top three samples) and RORD-Val (the bottom two samples).} Our OmniEraser demonstrates superior capability in removing objects and effects and preserving background.}
    \vspace{-0.5em}
    \label{fig:vis_main}
\end{figure*}

\noindent \textbf{Evaluation Datasets. }
Previous works \cite{jiang2025smarteraser,sagong2022rord,Li2025RORemTA} use the RORD validation dataset \cite{sagong2022rord} to evaluate their performance in real-world scenarios. However, the input masks in this dataset include regions affected by the object, which can not evaluate a model’s ability to effectively remove both the object and its associated effects based solely on the object itself. To address this gap, we introduce a new benchmark, \textit{RemovalBench}, as shown in Fig. \ref{fig:removalbench}. We capture each scene both with and without the object using a fixed-view camera and manually generate the object mask using the Segment Anything Model (SAM) \cite{kirillov2023sam}. \textit{RemovalBench} consists of 70 samples, each with 1024$\times$1024 resolution. Both \textit{RemovalBench} and RORD-Val are used as our primary comparison benchmarks. Following previous work \cite{jiang2025smarteraser}, we select 344 images from the original RORD validation set to create RORD-Val, ensuring both unique scenes and objects.

\noindent \textbf{Evaluation Metrics. }
We consider two aspects: image generation quality, and masked region preservation. For image quality, we use Frechet Inception Distance (FID) \cite{heusel2017fid}, CLIP Maximum Mean Discrepancy (CMMD) \cite{jayasumana2024cmmd}, and Asthetic Score (AS) \cite{schuhmann2022laion}. For masked region preservation, we follow previous works \cite{ju2024brushnet} using Learned Perceptual Image Patch Similarity (LPIPS) \cite{zhang2018lpips}, DINO Distance (DINO) \cite{oquab2023dinov2}, and Peak Signal-to-Noise Ratio (PSNR) \cite{korhonen2012psnr}.

\subsection{Comparison with Existing Methods. }
\label{subsec:exp_compare}
\noindent \textbf{Quantitative Results. } 
As shown in Tab. \ref{tab:main_bench}, OmniEraser significantly outperforms other approaches across all metrics on both benchmarks. Specifically, on our \textit{RemovalBench}, OmniEraser surpasses the previous state-of-the-art from 55.49 to 39.52 in FID, from 0.146 to 0.133 in LPIPS, and 0.232 to 0.208 in CMMD. 
Furthermore, on RORD-Val \cite{sagong2022rord}, which contains more diverse scenes and objects, our method achieves the best values of 43.71 in FID, 0.153 in CMMD, 22.13 in PSNR, and 4.99 in AS. These results demonstrate that OmniEraser excels not only in background preservation but also in maintaining overall image quality.

\begin{figure}[]
    \centering
    \includegraphics[width=\columnwidth]{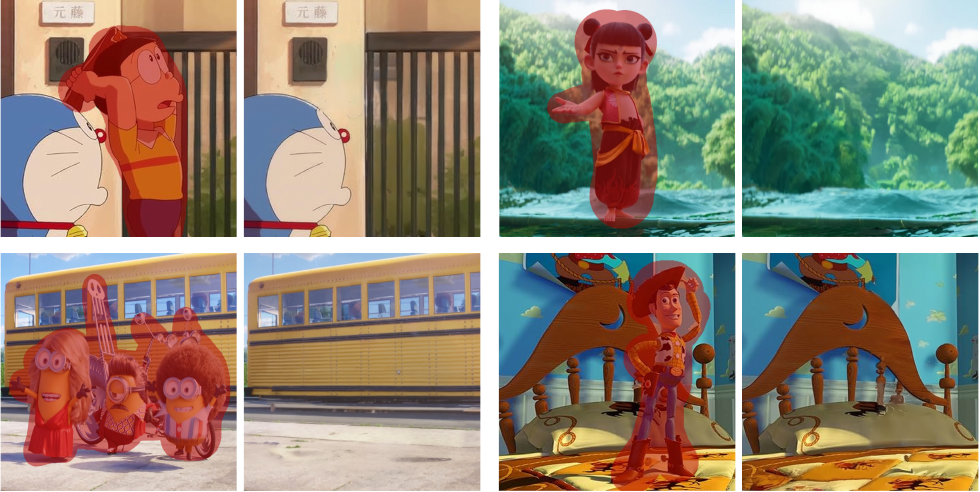}
    \caption{\textbf{Qualitative results on anime-style images.} Thanks to the powerful generative prior, OmniEraser exhibits strong capability despite not being trained on these samples.}
    \label{fig:vis_cartoon}
    \vspace{-1em}
\end{figure}

\begin{figure*}[t]
    \centering
    \includegraphics[width=\textwidth]{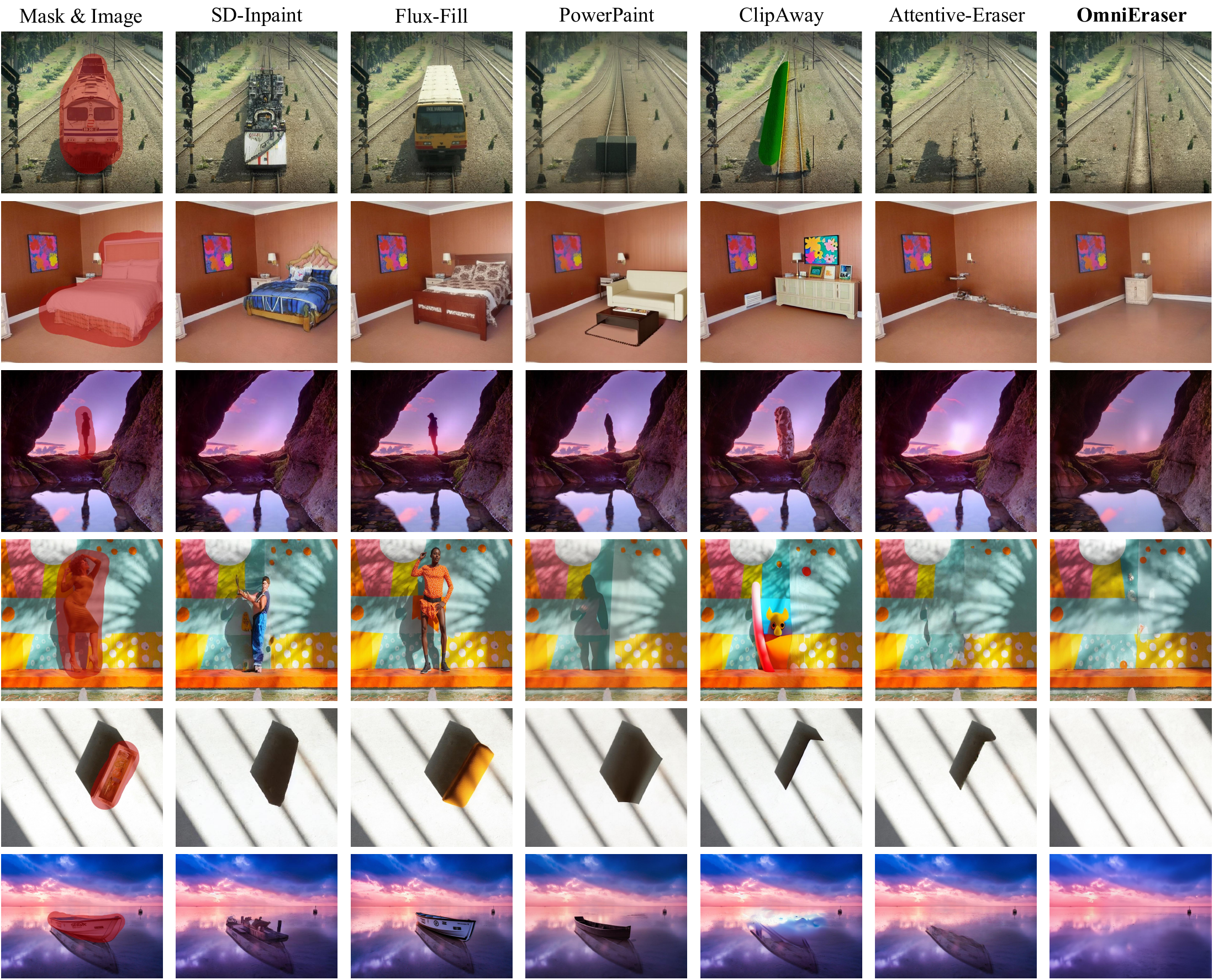}
    \vspace{-1.5em}
    \caption{\textbf{Qualitative results on in-the-wild cases.} Our OmniEraser effectively removes objects along with their shadows and reflections, demonstrating superior performance over previous approaches, particularly in complex and challenging scenes.}
    \vspace{-1.5em}
    \label{fig:vis_realcase}
\end{figure*}

\noindent \textbf{Qualitative Results. }
Fig. \ref{fig:vis_main} illustrates the visualization results of all approaches. Notably, the two benchmarks represent different settings: one where the masks include both the object and its associated effects (e.g., shadows and reflections), and another where the masks contain only the object itself. The results demonstrate the consistently superior performance of OmniEraser across both mask settings.

\noindent \textbf{In-the-wild Cases. }
To further evaluate our model in in-the-wild scenarios, we manually collect photorealistic images with prominent shadows and reflections from the web and the test set of the Emu Edit benchmark \cite{sheynin2024emu}. Some examples are illustrated in Fig. \ref{fig:vis_realcase}. We observe two main limitations in previous methods: 1) generate unintended content based on surrounding shadows, and 2) fail to erase the object's shadow. In contrast, our OmniEraser successfully removes both the object and its associated effects. 
Moreover, we present results from the anime-style images in Fig. \ref{fig:vis_cartoon}. Thanks to the strong generative capability of the base model we used, OmniEraser demonstrates impressive generalization, even though it was trained solely on real-world image data. Please refer to our Supplementary Material for more visualizations. We also compare OmniEraser with existing instruction-based image inpainting methods \cite{sheynin2024emu,fu2023guiding}, please refer to our Supplementary Material for details.

\begin{figure}[ht]
\centering
\includegraphics[width=\linewidth]{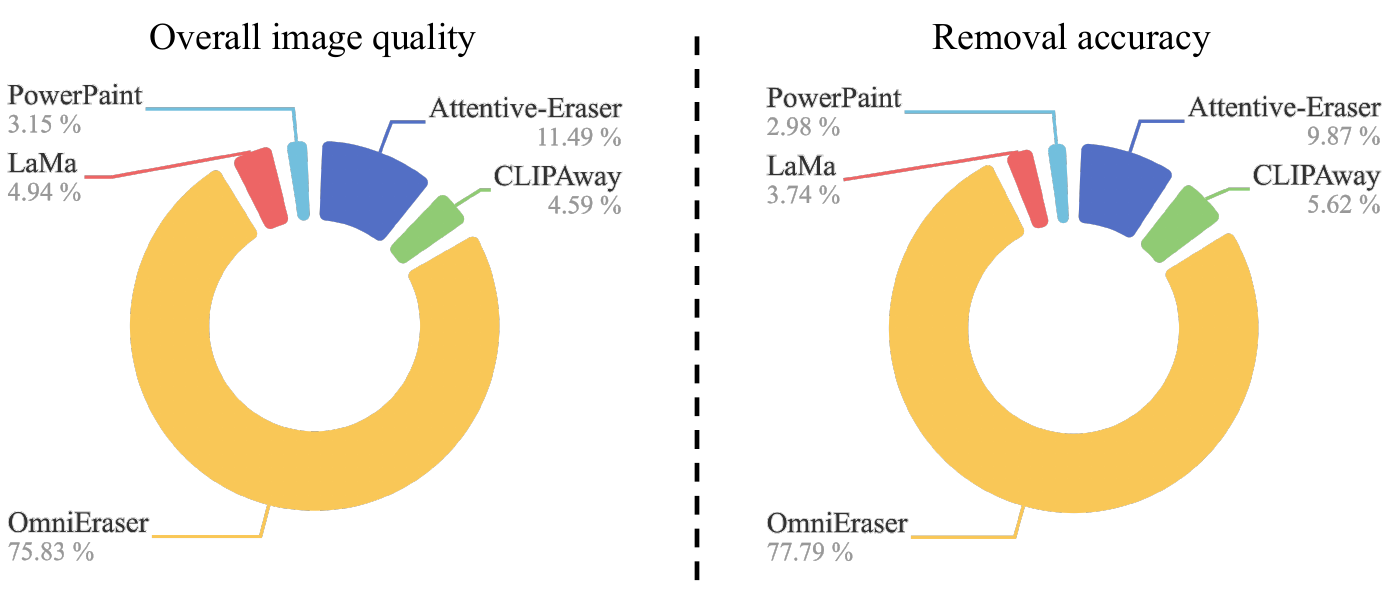}
\caption{\textbf{The results of our user study.} Our OmniEraser is preferred over previous methods across all criteria.}
\label{fig:user_study}
\vspace{-0.5em}
\end{figure}

\noindent \textbf{User Study.}
Following \cite{jiang2025smarteraser,yang2024generative}, we gather user preferences from 50 randomly selected participants using 55 generated examples. 
Each participant was asked to choose their preferred result from five different models, based on two criteria: overall image quality, and accuracy of objects and effects removal. As shown in Fig.~\ref{fig:user_study}, our method was preferred by more users across two criteria. 

\begin{figure*}[t]
    \centering
    \includegraphics[width=\textwidth]{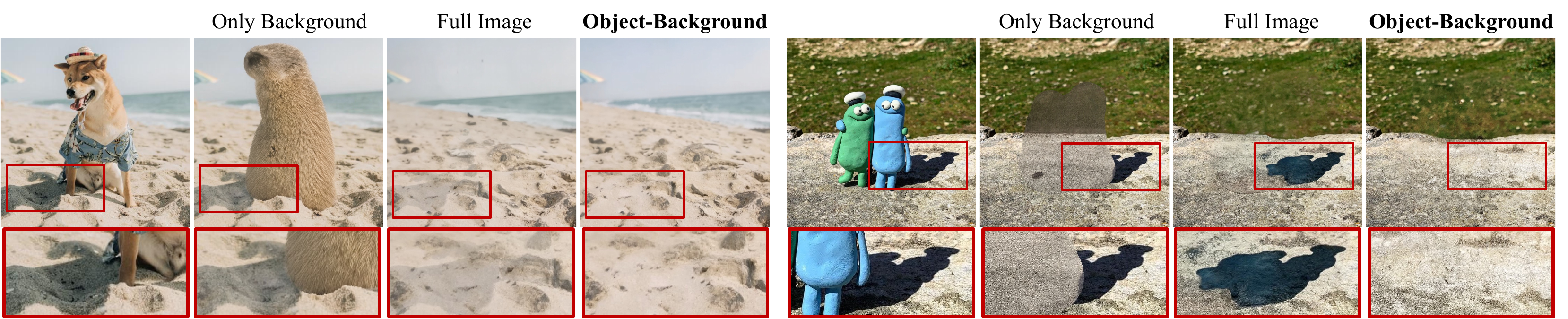}
    \caption{\textbf{Qualitative ablation of different guidance methods.} Using only the background and the full image generates artifacts and struggles to remove the object's shadow. Our Object-Background guidance can effectively remove both the object and its shadow. }
    \vspace{-0.5em}
    \label{fig:vis_guidance}
\end{figure*}

\subsection{Ablation Study and Analysis}
\label{subsec:exp_ablation}

\noindent \textbf{Comparison with Other Datasets. }
We compare our \textit{Video4Removal} dataset with other existing datasets, including RORD \cite{sagong2022rord} and MULAN \cite{tudosiu2024mulan}. MULAN \cite{tudosiu2024mulan} creates paired images for training object removal algorithms by decomposing images into multi-layer RGBA stacks. Each layer consists of an instance image with a transparent alpha channel and a background image, generated using off-the-shelf models. For a fair comparison, we use FLUX.1 as the prior model, keeping all parameters fixed, including training steps, seed, learning rate, and others. Results on our \textit{RemovalBench} are shown in Tab. \ref{tab:dataset}. Our \textit{Video4Removal} achieves the best performance, and we prove the importance of training on high-quality, real-world data.

\begin{table}[]
\renewcommand\arraystretch{1.15}
\centering
\resizebox{\columnwidth}{!}{%
\begin{tabular}{lcccc}
\toprule
Dataset & FID $\downarrow$ & CMMD $\downarrow$ & LPIPS $\downarrow$ & DINO $\downarrow$ \\ 
\midrule
MULAN \cite{tudosiu2024mulan} & 97.49 & 0.401 & 0.191 & 0.013 \\
RORD \cite{sagong2022rord} & 85.27 & 0.621 & 0.209 & 0.014 \\
\cellcolor{mycolor_green} \textit{Video4Removal} & \cellcolor{mycolor_green} \textbf{39.52} & \cellcolor{mycolor_green} \textbf{0.208} & \cellcolor{mycolor_green} \textbf{0.133} & \cellcolor{mycolor_green} \textbf{0.007} \\ 
\bottomrule
\end{tabular}}
\caption{\textbf{Quantitative results of different training datasets.} The best performance is bold.}
\label{tab:dataset}
\vspace{-0.5em}
\end{table}

\noindent \textbf{Other Generative Priors. }
We also demonstrate the effectiveness of our designs across different text-to-image diffusion architectures, such as SDXL, which uses convolutional U-Net \cite{ronneberger2015u}, and SD3 \cite{peebles2023scalable}, which employs transformer. The results presented in the last two rows of Tab. \ref{tab:main_bench} show that our variants also achieve competitive performance.

\begin{table}[]
\renewcommand\arraystretch{1.15}
\centering
\resizebox{\columnwidth}{!}{%
\begin{tabular}{lcccc}
\toprule
 & FID $\downarrow$ & CMMD $\downarrow$ & LPIPS $\downarrow$ & DINO $\downarrow$ \\ 
\midrule
Only Background & 119.94 & 0.509 & 0.241 & 0.017 \\
Full Image & 95.28 & 0.495 & 0.231 & 0.016 \\
\rowcolor{mycolor_green} Object-Background & \textbf{39.52} & \textbf{0.208} & \textbf{0.133} & \textbf{0.007} \\ 
\bottomrule
\end{tabular}}
\caption{\textbf{Quantitative results of different guidance methods.} Our Object-Background Guidance outperforms others.}
\label{tab:guidance}
\vspace{-0.5em}
\end{table}

\noindent \textbf{Effectiveness of Object-Background Guidance.}
Our OmniEraser method leverages the foreground object and the background as independent conditions of the diffusion model, as illustrated in Fig. \ref{fig:method}. We compare it with two baselines: using only the background image (Only Background) and using the full image (Full Image), which are common practices in previous works. 
We test their performances using \textit{RemovalBench} in Tab. \ref{tab:guidance}. Quantitative results demonstrate that our approach significantly outperforms the baselines in both visual quality and removal accuracy. 
Qualitative results are shown in Fig. \ref{fig:vis_guidance}. It is clear that using only the background image as guidance tends to generate shape-like artifacts, as the model only focuses on the background context. Using the full image successfully removes the target object but fails to remove its shadow. In contrast, our method, which utilizes the object and background separately, can remove both the object and its associated effects.

\begin{table}[]
\renewcommand\arraystretch{1.15}
\centering
\resizebox{\columnwidth}{!}{%
\begin{tabular}{ccccccc}
\toprule
\multicolumn{3}{c}{Mask Augmentations} & \multirow{2}{*}{FID $\downarrow$} & \multirow{2}{*}{CMMD $\downarrow$} & \multirow{2}{*}{LPIPS $\downarrow$} & \multirow{2}{*}{DINO $\downarrow$} \\ 
Dilation & Erosion & Box &  &  &  &  \\
\midrule
 &  &  & 48.53 & 0.394 & 0.218 & 0.007 \\
\checkmark &  &  & 42.46 & 0.366 & 0.213 & 0.007  \\
\checkmark & \checkmark &  & 40.46 & 0.363 & 0.209 & 0.007 \\
\checkmark & \checkmark & \checkmark & \textbf{39.40} & \textbf{0.347} & \textbf{0.203} & 0.007 \\
\bottomrule
\end{tabular}}
\caption{\textbf{Ablation results of mask augmentation.} Applying all strategies achieves the greatest robustness to irregular masks. }
\label{tab:mask_aug}
\vspace{-1.5em}
\end{table}

\noindent \textbf{Effectiveness of Mask Augmentation. }
Tab. \ref{tab:mask_aug} presents the ablation study on our mask augmentation. To comprehensively evaluate the model’s robustness to various user inputs, we manually perturb the object masks in \textit{RemovalBench}, creating test masks that are either incomplete or excessively covered. It can be seen that adding all strategies in training yields the best results. Please refer to our Supplementary Material for qualitative results.

%% file: sec/5_conclusion.tex
\section{Conclusion}
In this paper, we present \textbf{OmniEraser}, a state-of-the-art object removal method that effectively removes objects and their visual effects. 
To achieve this, we propose a new large-scale dataset, \textit{Video4Removal}, which contains over 100,000 high-quality image pairs extracted from video frames.
To prevent generating unintended new objects and artifacts, we propose Object-Background Guidance, which takes the object and background images as independent guidance to the diffusion model. 
We also provide a new benchmark, \textit{RemovalBench} for more comprehensive evaluation. Experiments and a user study demonstrate that OmniEraser significantly outperforms previous methods.